\documentclass[preprint,12pt]{elsarticle}
\usepackage{graphicx}
\usepackage{lipsum}
\usepackage{color}
\usepackage{bm}

\usepackage{amsthm}
\usepackage{amsmath}
\usepackage{amssymb}

\usepackage{amstext} 
\usepackage{array}   
\newcolumntype{L}{>{$}l<{$}} 

\theoremstyle{plain}
\newtheorem{thm}{Theorem}[]

\theoremstyle{definition}
\newtheorem{defn}{Definition}[]

\theoremstyle{remark}
\newtheorem{oss}{Remark}[]

\renewcommand{\vec}[1]{\bm{#1}}

\newcommand{\matr}[1]{\bm{#1}}     

\usepackage[notref, notcite]{showkeys}

\usepackage[nolist]{acronym}

\usepackage{hyperref}

\title{Learn to Synchronize, Synchronize to Learn}
\date{\today}

\author[1]{Pietro Verzelli\corref{mycorrespondingauthor}}
\cortext[mycorrespondingauthor]{Corresponding author: verzep@usi.ch}
\author[1,2]{Cesare Alippi}
\author[3,4]{Lorenzo Livi}

\address[1]{Faculty of Informatics, Università della Svizzera Italiana, Lugano, 69000, Switzerland.}
\address[2]{Department of Electronics, Information and bioengineering, Politecnico di Milano, Milan, 20133, Italy.}
\address[3]{Departments of Computer Science and Mathematics, University of Manitoba, Winnipeg, MB R3T 2N2, Canada.}
\address[4]{Department of Computer Science, College of Engineering, Mathematics and Physical Sciences, University of Exeter, Exeter EX4 4QF, United Kingdom.}

\journal{ArXiv}

\begin{document}

\begin{frontmatter}

\begin{abstract}
In recent years, the machine learning community has seen a continuous growing interest in research aimed at investigating dynamical aspects of both training procedures and machine learning models.
Of particular interest among recurrent neural networks we have the Reservoir Computing (RC) paradigm characterized by conceptual simplicity and a fast training scheme. 
Yet, the guiding principles under which RC operates are only partially understood.
In this work, we analyze the role played by Generalized Synchronization (GS) when training a RC to solve a generic task. 
In particular, we show how GS allows the reservoir to correctly encode the system generating the input signal into its dynamics. 
We also discuss necessary and sufficient conditions for the learning to be feasible in this approach.
Moreover, we explore the role that ergodicity plays in this process, showing how its presence allows the learning outcome to apply to multiple input trajectories.
Finally, we show that satisfaction of the GS can be measured by means of the Mutual False Nearest Neighbors index, which makes effective to practitioners theoretical derivations.
%
\end{abstract}

\end{frontmatter}

\section{Introduction}

The scientific community has seen a rising interest in research aimed at coupling machine learning and dynamical systems.
In fact, recent investigations have shown how  the theory developed for dynamical systems was useful to understand machine learning algorithms \cite{liu2019deep,bianchi2016investigating,sussillo2009generating, bengio1994learning};
the opposite holds, e.g. see \cite{bouvrie2017kernel, qi2020using, gilpin2020deep, tu2013dynamic, berry2020bridging}.

Within machine learning the \ac{RC} paradigm \cite{verstraeten2007experimental,lu2017reservoir} is particularly appealing due to its simplicity, cheap training mechanism and  state-of-the-art results obtained in solving various tasks \cite{lu2018attractor, chattopadhyay2020data, vlachas2020backpropagation, bompas2020accuracy}.
\ac{RC} was introduced independently by \citet{jaeger2001echo} (who used the term Echo State Network), \citet{maass2002real} (Liquid State Machine) and \citet{tino2001predicting} (Fractal Predicting Machine).
In order to account for a Neural Network implementation of \ac{RC} we use the term \ac{RCN} in the sequel.
The working principle of \ac{RC} relies on creating a representation of the input sequence by feeding it to an untrained dynamical system, \emph{the reservoir}, which should encode all relevant dynamics associated with the input. 
Learning focuses solely on the \emph{readout} function, which is trained to generate the desired output, given the encoded dynamics and the task at hand.

Some recent efforts have been devoted to understanding the encoding and learning mechanisms of \ac{RC} and their
capability to approximate dynamical systems. 
In particular, it was proven that \acs{RCN} are universal function approximators \cite{grigoryeva2018echo} and that their representations are rich enough to correctly embed dynamical systems through their state-space representation \cite{hart2019embedding, hart2020echo}.
Theoretical analysis of this learning principle led to many results about their expressive power \cite{gonon2020memory, massar2013mean, mastrogiuseppe2019geometrical, rivkind2017local, verstraeten2010memory}. 
Moreover, interesting results can be derived when assuming linear dynamics \cite{goudarzi2016memory, marzen2017difference, tino2020dynamical, verzelli2020input,ganguli2008memory}.
Due to its simple training mechanism, \ac{RC} is also particularly appealing for neuromorphic computing 
and other hardware implementations;
see \cite{TANAKA2019100} for a recent review.


The \ac{ESP} was introduced in the seminal work by \citeauthor{jaeger2001echo}~\cite{jaeger2001echo} as a necessary property for an effective and reliable computation.  
Basically, \ac{ESP} consists in requiring that the reservoir state  asymptotically depends only on the received input (i.e., the reservoir state \emph{echoes} the input) and does not depend on initial conditions of the reservoir.
Notably, even though most theoretical results assume the \ac{ESP} to hold \cite{grigoryeva2018echo,hart2019embedding}, existing sufficient conditions are too restrictive \cite{yildiz2012re} to be used in practical applications and necessary ones seem to suffice in most cases \cite{zhang2011nonlinear, basterrech2017empirical}.
In practice, some less restrictive criteria to verify satisfaction of the \ac{ESP} have been proposed over time \cite{yildiz2012re, manjunath2013echo,caluwaerts2013spectral, verstraeten2007experimental} as well as a general formulation for the \ac{ESP} accounting for multiple, stable responses to a driving input sequence \cite{ceni2020nESP}.
Yet, the problem with the \ac{ESP} verification lies on the fact that the \ac{ESP} definition does not explicitly take into account the structure of the driving input, which is simply defined as a sequence of values in an admissible range.
As a consequence, satisfaction of ESP cannot be verified but in simple cases for which the mathematics is amenable.

In order to verify the \ac{ESP}, we propose a new method based on a synchronization between dynamical systems.
In recent years, the concept of synchronization has been applied to \ac{RC} and yielded interesting results \cite{ lu2020invertible, weng2019synchronization, lymburn2019reservoir, grigoryeva2020chaos}.
The possibility of generalizing the concept of synchronization was first investigated by \citet{afraimovich1986stochastic} and \citet{rulkov1995generalized}, who introduced the term \ac{GS}. 
Successively, different empirical methods for verifying the presence of \ac{GS} from data have been introduced \cite{pecora1997fundamentals, parlitz2012detecting}.
A review on synchronization between dynamical systems was recently published~\cite{boccaletti2002synchronization}.

Recently, the \ac{GS} was compared to the \ac{ESP} and proposed as the basic working principle of \ac{RC} \cite{lu2018attractor,lu2020invertible,weng2019synchronization,grigoryeva2020chaos}.
In particular, under the assumption that there exists a dynamical system (called a \emph{source system}) generating the input data, \emph{the \ac{ESP} for the reservoir w.r.t. a driving input sequence is equivalent to the \ac{GS} between the reservoir and the (unknown) source system, with an additional requirement of uniqueness \cite{grigoryeva2020chaos}.}
This implies the existence of a stable \emph{synchronization manifold} to which the reservoir and the source system converge, and of a \emph{synchronization function} mapping states of the latter to states of the former.

In this work, we build on the seminal ideas developed in \cite{lu2018attractor,lu2020invertible} and discuss a novel methodology for dealing with a generic task. In particular, we focus on the implication that \ac{GS} has on the learning mechanism of \ac{RC}.
The scope of this work is two-fold: we aim at properly characterizing the equivalence between \ac{GS} and \ac{ESP} to show why \ac{GS} is needed in order for the \ac{RC} paradigm to work, and use these facts to interpret the \ac{RC} functionality under a new light.
More specifically, the novel aspects of this work can be summarized as follows: we show that when \ac{GS} occurs, the reservoir training may be viewed as a nonlinear basis expansion of the (unknown) source system state. 
This interpretation leads us to the development of two theorems, providing necessary and sufficient conditions for the learning to be realizable in the \ac{RC}-framework.
Moreover, we show that the ergodicity of the source system leads to the applicability of the learning methods to a general set of trajectories.
We then relax the realizability assumption and discuss why the \ac{GS} is necessary in that situation for the learning to happen.
Finally, we show how \ac{GS} can be easily verified for an \ac{RCN} driven by an input sequence, thus allowing one to assess the degree to which \ac{GS} holds for a specific input sequence driving the dynamics. For this we use an index, called the \ac{MFNN}, and empirically show that it is correlated with the \ac{RC} performance on the tasks at hand.

The paper is organized as follows: in Sec.~\ref{sec:RC} we introduce the theoretical framework, discussing the task we aim at solving and how this can be done with \ac{RC}. 
In Sec.~\ref{sec:GS} we present the concept of synchronization for dynamical systems and formalize the similarities with the concept of \ac{ESP}.
Sec.~\ref{sec:implications} contains the novel theoretical contributions of this paper and in Sec.~\ref{sec:experiments} we carry out simulations to validate the developed theory.
Finally, we draw conclusions in Sec.~\ref{sec:conclusions}.
The paper contains five appendices located at the end of this manuscript.

\section{Reservoir computing}
\label{sec:RC}
\begin{figure*}[htp!]
    \centering
    \includegraphics[width = \textwidth]{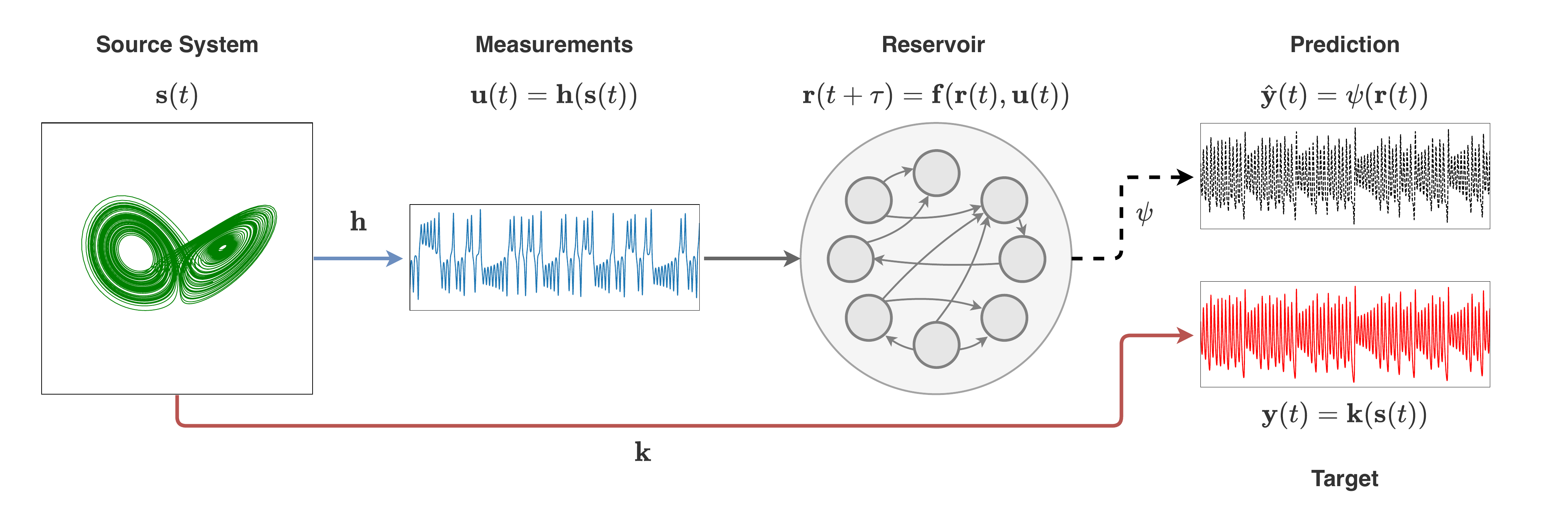}
    \caption{
    Diagram representing the \ac{RC} framework described in Section~\ref{sec:RC}. The source system $\vec{s}(t)$ evolves autonomously and generates the targets $\vec{y}(t)$ and the input measurements $\vec{u}(t)$.
    The latter is coupled to the reservoir $\vec{r}(t)$ so that its dynamics are dependent on (i.e., driven by) $\vec{u}(t)$. The readout $\vec{\psi}$ is then trained to generate the prediction $\hat{\vec{y}}(t)$, which is an approximation of $\vec{y}(t)$.}
    \label{fig:listening}
\end{figure*}

In this section we introduce the \ac{RC} setup by adopting the terminology used in \cite{lu2017reservoir}; the system formalization is general, independent of  the particular form of the source system or the reservoir.
A schematic representation of this approach is depicted in Fig.~\ref{fig:listening}.

\subsection{Task description}

Let us consider a discrete-time autonomous, noise-free \emph{source} system described by:
\begin{equation} \label{eqn:source_system}
    \vec{s}(t + \tau)= \vec{g}(\vec{s}(t))
\end{equation}
where $\vec{s}(t) \in \mathbb{R}^{d_{s}}$ denotes ${d_{s}}$-dimensional system \emph{state} at time $t$ and $\tau$ is the \emph{time increment}.
The \emph{source system} generates the time series to be exploited by the \ac{RC} architecture to solve a learning task.
Assume $\vec{g}$ to be differentiable and invertible, and that $\vec{s}(t)$ asymptotically approaches and stays in a bounded attractor, $\mathcal{A}_s$.
We are interested in the situation where $\vec{g}$ is unknown and we do not have direct access to the source system states.

The source system \eqref{eqn:source_system} produces two outputs, namely $\vec{u}(t) \in \mathbb{R}^{d_{u}}$ and $\vec{y}(t) \in \mathbb{R}^{d_{y}}$:
\begin{subequations}
    \begin{align}
    \vec{u}(t) = & \vec{h}(\vec{s}(t)) \label{eqn:measurement}\\
    \vec{y}(t) = & \vec{k}(\vec{s}(t)) \label{eqn:target}
    \end{align}
\end{subequations}
We name $\vec{u}$ as the \emph{measurements} (or \emph{observables}), i.e., the available input data.
The vector-valued function $\vec{h}(\cdot)$ is introduced to account for the fact that a function of $\vec{s}$ is used to generate the data.
We refer to $\vec{y}$ as the \emph{targets}, i.e., the supervised information describing the task to be learned. The targets are generated via a vector-valued function $\vec{k}(\cdot)$.
Both $\vec{h}(\cdot)$ and $\vec{k}(\cdot)$ are unknown.
We assume that there is no measurement noise, as commonly done in the related literature \cite{lu2017reservoir, lu2017reservoir, hart2019embedding, hart2020echo}.

\begin{figure}[ht!]
    \centering
    \includegraphics[width =.8 \textwidth]{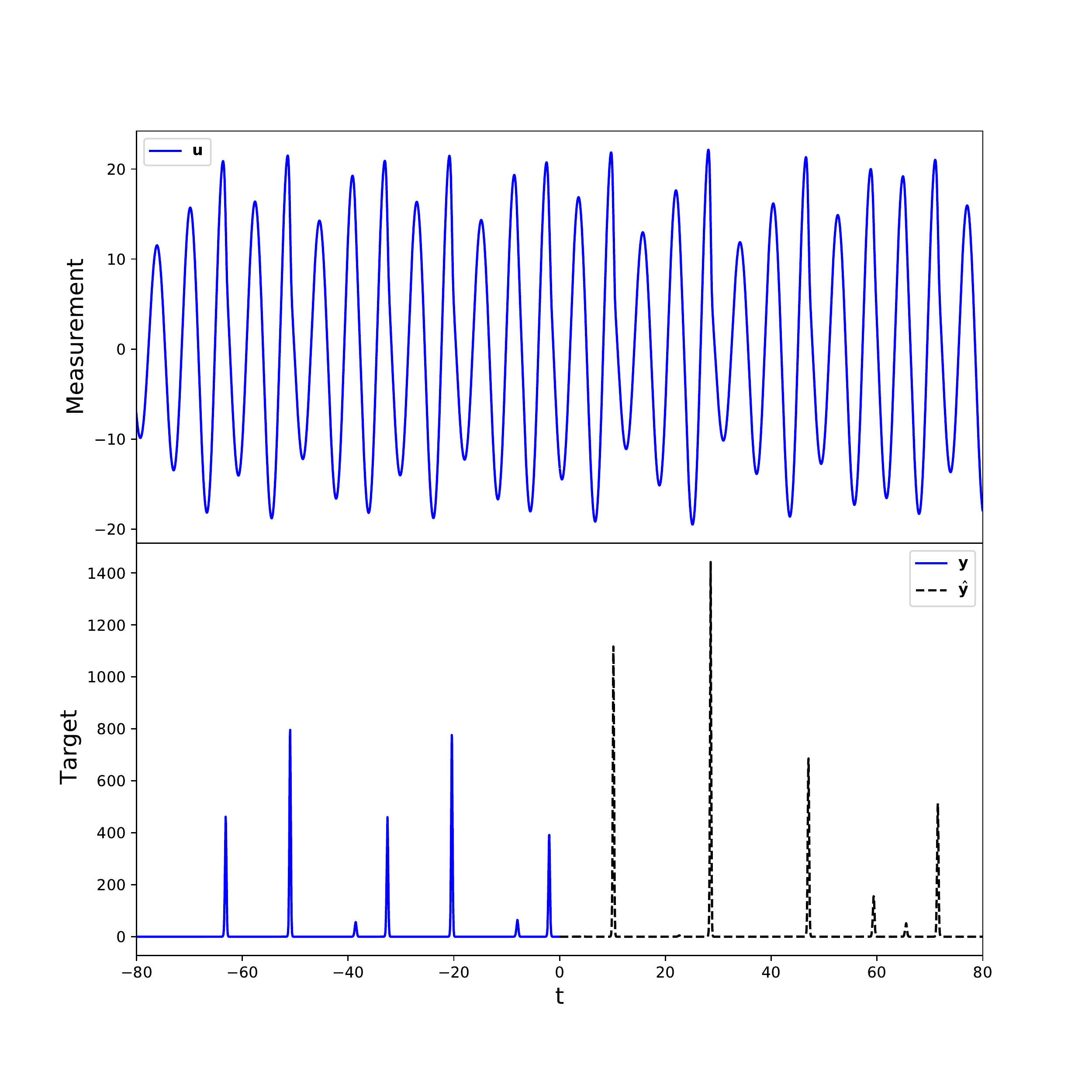}
    \caption{An example of the problem under study, where both $\vec{u}$ and $\vec{y}$ are mono-dimensional. 
    The input value $\vec{u}$ is always provided (top figure), while the target $\vec{y}$ is only accessible at training time, i.e., for $t<0$ (bottom figure, blue solid line).
    The goal is to generate a prediction $\hat{\vec{y}}$ for $t>0$ by using the input only.
    Here, the source system is the R{\"o}ssler system (see Appendix~\ref{sec:roessler}), the input is $u(t) = x(t)$ while the target is $y(t) = z^{2}(t)$, where $x(t)$ and $z(t)$ are two variables constituting the R{\"o}ssler system.}
    \label{fig:task_example}
\end{figure}

Both $\vec{u}(t)$ and $\vec{y}(t)$ are accessible for $t<0$ (\emph{training phase)}, but for $t \ge 0$ only $\vec{u}(t)$ is available.
Our goal is then to use the continued knowledge about $\vec{u}$ to generate a valid prediction $\hat{\vec{y}}(t)$ of $\vec{y}(t)$, for $t \ge 0$.
We call this phase the \emph{predicting phase}.\footnote{We choose this term -- following \cite{lu2017reservoir} -- to avoid the possible ambiguity between the testing and validation phases typically used in machine learning tasks, since this distinction is not well-defined in this context.}
Figure~\ref{fig:task_example} provides an example of the framework taken into account.

A typical instance of this problem is the forecasting task, say to predict the value $\vec{u}(t)$ will assume $d$ times ahead, hence providing $\vec{y}(t) = \vec{u}(t+d)$.
Another relevant task (called the \emph{observer task} \cite{lu2017reservoir}) requires to estimate the state of the system having information about $\vec{u}(t)$ only, i.e.,$\vec{y}(t) = \vec{s}(t)$.
An example of the framework is provided in Fig.~\ref{fig:task_example}.

\subsection{Training phase}

For the training phase, we assume to have access to a (possibly infinite) series of measurements $\vec{u}(t)$ and a paired series of target values $\vec{y}(t)$.
The goal of the training phase is to produce a function which generates an accurate prediction $\hat{\vec{y}}(t)$ of the target when reading $\vec{u}(t)$.
The problem lies on the fact that the target values $\vec{y}$ depend on the full state of the source $\vec{s}$, while only the measurements $\vec{u}$ are accessible. 
So, one needs to be able to represent the full state of the source from the measurements only and then use it to estimate the target function.
In the \ac{RC} approach, these two parts are explicitly separated.
To represent the full state, one uses a different dynamical system, \emph{the reservoir}, which creates a meaningful representation of the source system $\vec{s}$ when driven by the measurements $\vec{u}$.
We will call this part the \emph{listening} phase.
Then, a function must be used to compute the desired output from the reservoir states.
This is done by estimating a \emph{readout function}, which takes a reservoir state as input and produces an output. This phase is called the \emph{fitting} phase.

\paragraph{Listening}
In the \emph{listening} phase, the training measurements are used as input to the reservoir, which is modelled as a discrete-time%
\footnote{In fact, the theory also applies to continuous-time models. In \cite{lu2018attractor} the authors discuss the theory for discrete-time systems, but then use a continuous-time model in the experimental section.} %
deterministic driven dynamical system:
\begin{equation} \label{eqn:listening}
    \vec{r}(t + \tau) = \vec{f}(\vec{r}(t), \vec{u}(t+\tau))
\end{equation}

Here $\vec{r}(t) \in \mathbb{R}^{d_r}$ is the \emph{reservoir state}.
We assume $\vec{f}$ to be a differentiable function controlling the reservoir evolution. 

\paragraph{Fitting}
\emph{Fitting} consists in determining the so-called \emph{readout function}, $\vec{\psi}_{\vec{\theta}}$, which reads the reservoir state $\vec{r}(t)$ and provide an estimate for the output $\vec{y}(t)$.
The parameters $\vec{\theta}$ are selected by a fitting procedure yielding a parameter configuration $\hat{\vec{\theta}}$ such that:
\begin{equation}\label{eqn:training}
      \hat{\vec{y}}(t) = \vec{\psi}_{\hat{\vec{\theta}}}(\vec{r}(t))  \approx \vec{y}(t)
\end{equation}{}

From now on we will drop the $\hat{\vec{\theta}}$-notation and simply write $\vec{\psi}$ for $\vec{\psi}_{\hat{\vec{\theta}}}$.

It is important to point out that this is an offline learning procedure: the enriched input representation is first created through the reservoir states and only then the readout function is computed in one shot. 
Online learning procedures may be exploited as well, e.g. see \cite{sussillo2009generating,lu2020invertible}.

\subsection{Predicting  phase}

After training is complete, the system will be used to predict new target values (\emph{predicting phase}).%
It follows that, in this phase the reservoir is driven by $\vec{u}(t)$ and the new output is  $\hat{\vec{y}}(t)$. 
Being the readout time-independent, and in absence of an output feedback mechanism, the reservoir is subject to the \emph{same} dynamics during training and predicting phases, since the network continues to be driven by $\vec{u}$.%
\footnote{We implicitly assume that $\vec{u}$ is characterized by the same dynamics in both phases, implying some form a stationarity of the source system. Otherwise, the learning would be unfeasible without proper adaptation mechanisms to changes in the driving input. Namely, we are assuming that the source system \eqref{eqn:source_system} has reached its attractor in the listening phase and that it will continue to stay on it.}
In other words, the dynamical part of the system does not ``perceive'' in any way the change between the listening and the predicting phase: this is why the \ac{ESP} was introduced in this setting.


\section{Synchronization and echo-state property} \label{sec:GS}

\subsection{Drive-response systems}
In this section we introduce the concept of \ac{GS} and relate it to the concept of \ac{ESP}.
To do so, we start by considering the source system \eqref{eqn:source_system} together with the reservoir \eqref{eqn:listening}  in a \emph{drive-response system}: 
\begin{subequations} \label{eqn:drive_response}
    \begin{align}
        \vec{s}(t+\tau) &= \vec{g}(\vec{s}(t))  &\text{(Drive)} \label{eqn:drive}\\
        \vec{r}(t+\tau) &= \vec{f}(\vec{r}(t), \vec{u}((t+\tau)))
        =  \vec{f}(\vec{r}(t), \vec{h}(\vec{s}(t+\tau))) &\text{(Response)} \label{eqn:response}
    \end{align}
\end{subequations}

Where \eqref{eqn:drive} is the \emph{drive} and  \eqref{eqn:response} the \emph{response}. 
Together, they form an autonomous ($d_s + d_r$)-dimensional dynamical system which can be written as:
\begin{equation}\label{eqn:original}
    \vec{x}(t+\tau) = \vec{G}(\vec{x}(t))  
\end{equation}
where $\vec{x}$ is simply the concatenation of $\vec{s}$ and $\vec{r}$ and, accordingly, $\vec{G}$ represents the concatenation of the action of $\vec{g}$ and $\vec{f}$.

Let us now assume that \eqref{eqn:original} has an attractor $\mathcal{A}$ with a basin of attraction $\mathcal{B}$.%
\footnote{This assumption is required just to simplify the exposition. For the case where the system has multiple attractors see, for instance, \cite{lu2020invertible}.}
This attractor can be expressed as:
$$
\mathcal{A} = \mathcal{A}_s \times \mathcal{A}_r
$$
where $\mathcal{A}_s$ (respectively, $\mathcal{A}_r$) is the projection of $\mathcal{A}$ into the $d_s$(respectively, $d_r$) coordinates of system $\eqref{eqn:drive}$ (respectively, $\eqref{eqn:response}$).
The same holds for the basin of attraction $\mathcal{B}$ of $\mathcal{A}$, which can be expressed in an analogous way as
$$
\mathcal{B} = \mathcal{B}_s \times \mathcal{B}_r
$$

It is important to note that, because \eqref{eqn:drive} is an \emph{autonomous dynamical system}, $\mathcal{A}_s$  is its attractor and $\mathcal{B}_s$ its basin of attraction. The nature of $\mathcal{A}_r$ and $\mathcal{B}_r$ is more complex, as \eqref{eqn:response} is a non-autonomous dynamical system, for which the definition of attractor is more complicated (and non-uniquely defined \cite{manjunath2012theory,ceni2020nESP}): for our purpose, $\mathcal{A}_r$ and $\mathcal{B}_r$ can be simply thought of as sets obtained by a projection of the whole system considering only the variables related to the response system.

\subsection{Generalized Synchronization}\label{subsec:GS}
As stated in the introduction, the concept of synchronization has been raising interest in the \ac{RC} community. 
Here we introduce the \ac{GS}, which is a generalization of the concept of synchronization for non-identical systems. A short introduction of the simpler case in which the synchronization occurs between identical system can be found in Appendix~\ref{sec:identical_sync}).
We introduce \ac{GS} following the definition used in \cite{parlitz2012detecting}:
\begin{defn}[Generalized synchronization]\label{def:GS}
A system like \eqref{eqn:drive_response} possesses the property of \acf{GS} when there exist a transformation 
\begin{align} \label{eqn:GS_func}
    \vec{\phi} : \quad  \mathbb{R}^{d_s} &\to \mathbb{R}^{d_r} \\
                 \vec{s} &\mapsto \vec{\phi}(\vec{s})
\end{align}
mapping the states of the drive into the states of the response for which:
\begin{equation} \label{eqn:GS_asymptotic}
    \lim_{t \to \infty} \lVert \vec{r}(t) - \vec{\phi}(\vec{s}(t)) \rVert = 0
\end{equation}
\end{defn}

This means that the response state $\vec{r}$ is asymptotically given by the state of the driving system $\vec{s}$ and there exists a \emph{synchronization manifold} $\mathcal{M}$ in the full state-space of the system defined by the equation:

\begin{equation} \label{eqn:GS_condition}
\vec{r} = \vec{\phi}(\vec{s}).
\end{equation}

i.e., $\mathcal{M}:= \{(\vec{s},\vec{r}) : \vec{r} = \vec{\phi}(\vec{s})\}$.
Clearly $\mathcal{M} \subseteq \mathcal{A}$. Moreover we can define $\mathcal{B}_\mathcal{M}$ as the set of initial conditions for which \eqref{eqn:GS_asymptotic} holds. Then $\mathcal{B}_\mathcal{M} \subseteq \mathcal{B}$.
As noted in \cite{rulkov1995generalized}, if a synchronizing relationship of the form \eqref{eqn:GS_condition} occurs, it means that the motion of the system in the full space has collapsed onto a subspace which is the manifold of the synchronized motion $\mathcal{M}$.
This manifold is invariant, in the sense that $\vec{r}(t) = \vec{\phi}(\vec{s}(t))$ implies $\vec{r}(t+\tau) = \vec{\phi}(\vec{s}(t+\tau))$.
Moreover, \eqref{eqn:GS_asymptotic} implies that such a manifold must be attracting \cite{parlitz2012detecting}.

Since the relationship defined in \eqref{eqn:GS_asymptotic} should hold on the attractor $\mathcal{A}_{s}$, which the drive system approaches asymptotically, it makes sense to write the attractor of the response system as  $\mathcal{A}_{r} = \vec{\phi}(\mathcal{A}_{s})$. 
We assume $\vec{\phi}$ to be smooth (which can be theoretically granted for a large class of systems \cite{grigoryeva2020chaos}).  The case in which the synchronization function exists but is complicated or even fractal is called \emph{Weak Synchronization} \cite{pyragas1996weak}; this case is not taken into account in our paper. If $\vec{\phi}$ equals the identity transformation, this general definition of synchronization coincides with the definition of identical synchronization (see Appendix \ref{sec:identical_sync}).

\subsection{Echo State Property}

The motivation behind the original \ac{ESP} formulation is the following: for learning to be realizable, it is crucial that the current network state $\vec{r}(t)$ is uniquely determined by the input sequence $\{\vec{u}(t)\}$.
Such a definition takes into account a specific input sequence, with values in a compact set $\mathcal{U}$; in practical applications, the input will always be bounded. Also the compactness of the reservoir state-space is required, but it is automatically guaranteed if one considers a bounded nonlinear activation functions (like $\tanh$).
\begin{defn}[Compatibility]\label{def:compatibility}
We say that a state sequence $\{ \vec{r}(t) \}$ is \emph{compatible} with a bounded input sequence $\{\vec{u}(t)\}$ when, for all $t$: 
\[
\vec{r}(t+\tau) = \vec{f}(\vec{r}(t), \vec{u}(t+\tau)) 
\]
\end{defn}{}


\begin{defn}[ESP \cite{jaeger2001echo}]
The system has the \acf{ESP} if for every input sequence $\{\vec{u}(t)\}$, for any state sequences $\{ \vec{r}_1(t) \}$ and  $\{ \vec{r}_2(t) \}$ compatible with $\{\vec{u}(t)\}$ it holds that $\vec{r}_1(t) = \vec{r}_2(t)$ for each $t$. 
\end{defn}

This means that a state $\vec{r}(t)$ is uniquely determined by any left-infinite input sequence. This can be stated in an equivalent way by requiring the existence of a \emph{input echo function}
$\vec{E} = (e_1, \dots, e_{d_r})$ where $e_i:\mathcal{U}^{-\mathbb{N}} \to \mathbb{R}$ such that for all left-infinite input histories $\dots, \vec{u}(t-1), \vec{u}(t)$ the current state is given by:
$$\vec{r}(t) = \vec{E}(\dots, \vec{u}(t-1), \vec{u}(t))$$


The assumption that the input $\vec{u}(t)$ is given by \eqref{eqn:measurement} (i.e., it is a function of the state of an autonomous dynamical system) makes it possible to explore the matter in more dept.
We start by noticing that, in this framework, a sequence $\{u(t)\}$ is uniquely defined by an initial condition of \eqref{eqn:source_system} as:
$$
\vec{u}(t) = \vec{h}(\vec{s}(t) = \vec{h}(\vec{g}^t({\vec{s}_0}))
$$
which holds for any $t$ as $\vec{g}$ is assumed to be invertible.
This means that each left-infinite sequence of measurements can be uniquely associated to a state of the source system \eqref{eqn:source_system} so that:
$$
\vec{E}(\dots, \vec{u}(t-1), \vec{u}(t)) = \vec{E}(\dots, \vec{h} 
\circ \vec{g}^{-1} \circ \vec{s}(t), \vec{h} \circ \vec{s}(t) )
$$
which is clearly a function of $\vec{s}(t)$ only and is, in fact, \eqref{eqn:GS_condition}.
Within our framework, the existence of an input echo function is equivalent to the existence of a synchronization function, i.e.:
$$
\vec{E}(\dots, \vec{u}(t-1), \vec{u}(t)) = \vec{\phi}(\vec{\vec{s}(t)}) 
$$

Yet, the analogy is not perfect as the \ac{ESP} requires $\vec{\phi}$ to be \emph{unique}.
Non-uniqueness means that there exists $p>1$ different synchronization manifolds, each one given by a different synchronization function $\mathcal{M}^i:= \{(\vec{s},\vec{r}) : \vec{r} = \vec{\phi}^i(\vec{s})\}, i = 1,\dots,p$. In \cite{grigoryeva2020chaos} the authors show that this phenomenon can be avoided by ensuring local contractivity of $\vec{f}$, i.e. $\vec{f}$ should operate as a contraction on each separate manifold $\mathcal{M}^i$.

When evaluating the reservoir system performance (which is needed in order to perform hyper-parameters tuning), one usually compares single realizations of the reservoir and of the input signal, i.e. a specific instance of the reservoir with its initial condition is trained on an input signal. 
This means that the uniqueness is not practically exploited in most practical context, and sometimes it might even be detrimental (see the concept of ``echo index'' introduced in \cite{ceni2020nESP}): for this reason, in this paper we simply explore the \ac{GS} disregarding its uniqueness.

\section{Generalized synchronization and learning}
\label{sec:implications}

In the scenario depicted above, one uses the reservoir $\vec{r}$ to create a representation of the input, which is finally processed by the readout $\vec{\psi}$.
The goal is to generate a mapping from $\vec{s}$ to $\vec{y}$ and then to use such readout for generating $\hat{\vec{y}}(t)$ for values of $\vec{u}(t)$ which are not in the training set (i.e., for $t \ge 0$).
Since $\vec{s}$ is unknown, what one really assumes is that it is possible to predict $\vec{y}$ from the knowledge of the whole history of $\vec{u}$.
This is, in fact, an implication of \emph{Takens embedding theorem} \cite{takens1981detecting} and the feasibility of such a procedure was recently proved in the context of \ac{RC} by \citet{hart2019embedding}\footnote{Note that what they call \emph{echo state map} (see Theorem~2.2.2 in \cite{hart2019embedding}) corresponds to the synchronization function in \eqref{eqn:GS_condition}}.

It is really important to emphasize the fact that we only consider the case where the fitting of the readout \emph{does not} affect the reservoir dynamics in any way.
The representation of the attractor of $\vec{s}$ into the reservoir states $\vec{r}$ by the use of the input sequence $\vec{u}$ is done in the listening phase, which is (in machine learning parlance) \emph{unsupervised}.
The fitting consists of trying to estimate the \emph{static} function $\vec{k}$ mapping the state $\vec{s}$ to the desired output $\vec{y}$, i.e,
\begin{equation} \label{eqn:readout_goal}
 \hat{\vec{y}}(t) := \vec{\psi}(\vec{r}(t))\approx \vec{k}(\vec{s}(t)) = \vec{y}(t) 
 \qquad \forall t
\end{equation}

We now discuss the role that the listening phase has on the learning process.

\subsection{Unsupervised system reconstruction during the listening phase}

Let us consider the time interval $(t_s, 0)$, in which we assume that the \ac{GS} has occurred; remember that we assume negative times for the training phase, so $t_s<0$.
We consider the reservoir states generated in this interval,
\begin{multline}\label{eqn:R_matrix}
\matr{R}_{(t_s, 0)}=
\begin{bmatrix}
    \vert           &\vert                  &\vert      &\vert \\
    \vec{r}(t_s)    &\vec{r}(t_s + \tau)    &\dots      &\vec{r}(0)    \\
    \vert           &\vert                  &\vert      &\vert 
\end{bmatrix}
=\\=
\begin{bmatrix}
    \vert           &\vert                  &\vert      &\vert \\
    \vec{r}(t_s)    &\vec{f}(\vec{r}(t_s), \vec{u}(t_s))   &\dots      &\vec{f}(\vec{r}(-\tau), \vec{u}(-\tau)) \\
    \vert           &\vert                  &\vert      &\vert 
\end{bmatrix}
\end{multline}


\ac{GS} guarantees that there exists a function mapping the source system states to the reservoir states and also its invariance. This means that
\begin{equation} \label{eqn:implication_invariance}
    \vec{r}(t) = \vec{\phi}(\vec{s}(t)) \Rightarrow 
    \vec{r}(t+\tau) = \vec{\phi}(\vec{s}(t + \tau)) = \vec{\phi}(\vec{g}(\vec{s}(t)))
\end{equation}
so that \eqref{eqn:R_matrix} can be written as follows:
\begin{equation}\label{eqn:R_static}
\matr{R}_{(t_s, 0)}=
\begin{bmatrix}
    \vert           &\vert                  &\vert      &\vert \\
    \vec{\phi}(\vec{s}(t_s))    &\vec{\phi}(\vec{s}(t_s + \tau))     &\dots      &\vec{\phi}(\vec{s}(0))    \\
    \vert           &\vert                  &\vert      &\vert 
\end{bmatrix}
\end{equation}

Note that $\vec{\phi}$ is a time-independent function that is the same for all $\vec{s}$.
Since by assumption $d_r > d_s$, we can think of $\vec{\phi}$ as an attempt to expand the source system state-space (which is unknown) into a higher-dimensional space, in the same fashion as the well-known reproducing kernel Hilbert space mechanism behind kernel methods \cite{shawetaylor+cristianini2004}: the reservoir dynamics performs a sort of ``nonlinear basis expansion'' of the (unknown) attractor of $\vec{s}$. 
The use of the synchronization function $\vec{\phi}$ provides a sound theoretical framework to the fitting process, and the relation \eqref{eqn:GS_condition} can be seen a sound formulation of the ``reservoir trick''; see \cite{shi2007support}. 
Moreover note that such an expansion $\vec{\phi}$ was not computed or estimated from data, but was ``obtained'' as a result of driving the reservoir with the input sequence under consideration: this means that the mapping is ``informed'' of the dynamics.
Accordingly, we can interpret \eqref{eqn:readout_goal} as follows:
\begin{equation} \label{eqn:readout_true}
    \hat{\vec{y}}(t) = \vec{\psi}(\vec{r}(t)) = \vec{\psi}(\vec{\phi}(\vec{s}(t))) \approx \vec{k}(\vec{s}(t)) = \vec{y}(t)
\end{equation}

\subsection{Learning realizability}

We define the concept of ``realizable learning'' \cite{shalev2014understanding} as the situation where the readout is perfectly able to reconstruct the targets by using the reservoir states. More formally,
\begin{defn}[Learning realizability]
\label{def:realizability}
We say that the learning is \emph{realizable} if there exists a readout $\vec{\psi}$ such that, 
\begin{equation}\label{eqn:realizability}
\vec{y}(t) = \vec{\psi}(\vec{r}(t)),\ \forall t
\end{equation}
\end{defn}

The following theorem proves that for the learning to be realizable for a trajectory of the source system, there must be a function mapping that trajectory into the trajectory of the reservoir.
First we introduce some notation.
Let us denote with $\mathcal{S} \subset \mathcal{A}_{s}$ the set containing all $\vec{s}(t)$, for all $t$ (this is usually called the \emph{orbit} of a system).
Analogously, we define $\mathcal{R}\subset \mathcal{A}_{r}$ as the set of all $\vec{r}(t)$, for all $t$.
We define $\mathcal{Y}$ as the result of applying $\vec{k}$ to each point in $\mathcal{S}$, in short $\mathcal{Y} := \vec{k}(\mathcal{S})$.
\begin{thm}
A necessary condition for learning to be realizable is that for each $\vec{r} \in \mathcal{R}$ such that $\vec{\psi}(\vec{r}) = \vec{y}$ , there exists a function $\vec{\mathcal{F}}: \mathcal{S} \to \mathcal{R}$ such that  $\vec{r} = \vec{\mathcal{F}}(\vec{s})$, where $\vec{s}$ is such that by $\vec{k}(\vec{s}) = \vec{y}$.
\end{thm}

\begin{proof}
Realizability of learning implies that $\vec{\psi}$ is surjective when mapping $\mathcal{R}$ into $\mathcal{Y}$.
The surjectivity of $\vec{k}$ is guaranteed by the way we constructed $\mathcal{Y}$.
But because different source system states could result in the same target, $\vec{k}$ may not be an injective function. 
The same holds for $\vec{\psi}$.
We define $\vec{\psi}^{\dagger}(\vec{y})$ as a function mapping each $\vec{y}$ onto an $\vec{r}$:
if $\vec{\psi}$ is also injective, then $\vec{\psi}^{\dagger}$ corresponds to the inverse of $\vec{\psi}$, but in general it is not.
These functions are called \emph{right-inverse} since $\vec{\psi} \circ \vec{\psi}^{\dagger}$ is the identity but $\vec{\psi}^{\dagger} \circ \vec{\psi}$ is not.
Since by definition $\vec{y}(t) = \vec{k}(\vec{s}(t))$, it will then be possible to construct the function $\vec{\mathcal{F}}$ as follows:
\begin{equation} \label{eqn:F_construct}
    \vec{\mathcal{F}} = \vec{\psi}^{\dagger} \circ \vec{k}
\end{equation}

Such a function maps all $\vec{s}$ into the corresponding targets $\vec{y}$ and inverts the readout function $\vec{\psi}$ so that it maps each target to a corresponding reservoir state $\vec{r}$.

So far, we have shown that \eqref{eqn:F_construct} maps each $\vec{s}$ to an $\vec{r}$.
We also need to make sure that each $\vec{r}$ can be written as $\vec{\mathcal{F}}(\vec{s})$.
This is granted by our assumption (surjectivity of $\vec{\psi}$) which tells us that each $\vec{y}$ can be written as $\vec{\psi}(\vec{r})$. 
Then, using an argument analogous to the one above, we can associate each $\vec{y} \in \mathcal{Y}$ to an $\vec{s} \in \mathcal{S}$ by defining $\vec{k}^{\dagger}$. Again, this corresponds to the inverse of $\vec{k}$ only if $\vec{k}$ is also injective.
Note that, in general, distinct values of $\vec{r}$ might be associated to the same $\vec{s}$ (and \textit{viceversa}).
This shows that if learning is realizable, then $\vec{\mathcal{F}}$ must exist.
\end{proof}

The theorem also implies that if $\vec{\mathcal{F}}$ does not exist, then learning is not realizable.
So, any successful training procedure must (i) develop an (implicit) mapping from $\mathcal{S}$ to $\mathcal{R}$ and (ii) find a suitable readout.
Yet, the existence of $\vec{\mathcal{F}}$ \emph{does not necessarily imply} the realizability of learning: we have no guarantees that, in the presence of such a mapping, a readout solving the problem can be found.
Moreover, the fact that learning is not realizable \emph{does not necessarily imply} that $\vec{\mathcal{F}}$ does not exist: the problem might simply be that we are not able to conceive the right readout.

We now prove that, by requiring $\vec{\mathcal{F}}$ to be injective, we can always construct a readout which correctly solves the problem.
\begin{thm}\label{thm:sufficient_realizability}
A sufficient condition for the learning to be realizable is that there exists a function $\vec{\mathcal{F}}$ such that  for all $\vec{r} \in \mathcal{R}, \vec{r} = \vec{\mathcal{F}}(\vec{s})$ and $\vec{\mathcal{F}}$ is injective.
\end{thm}

Before proving the theorem, we make a remark:
\begin{oss} \label{oss:invertibility}
In Theorem \ref{thm:sufficient_realizability}, the condition that $\vec{r} = \vec{\mathcal{F}}(\vec{s})$ must hold for all $\vec{r} \in \mathcal{R}$ means that $\vec{\mathcal{F}}: \mathcal{S} \to \mathcal{R}$ must be surjective. This means that when $\vec{\mathcal{F}}$ is injective, it is in fact bijective and so, invertible.
\end{oss}

The proof of the theorem in now trivial:
\begin{proof}
As discussed in Remark \ref{oss:invertibility}, the injectivity of $\vec{\mathcal{F}}$ grants the existence of its inverse $\vec{\mathcal{F}}^{-1}$.
The readout function solving \eqref{eqn:realizability} then exists and it is given by:
\begin{equation} \label{eqn:psi_construction}
    \vec{\psi} = \vec{k} \circ \vec{\mathcal{F}}^{-1}
\end{equation}
\end{proof}

The fact that $\vec{\mathcal{F}}$ is injective means that it always maps distinct $\vec{s}$ into distinct $\vec{r}$. 
Without it, $\vec{\mathcal{F}}$ may map two distinct $\vec{s}_1, \vec{s}_2$ into the same $\vec{r} = \vec{\mathcal{F}}(\vec{s}_1) =\vec{\mathcal{F}}(\vec{s}_2)$: the realizability of learning then depends on whether  $\vec{k}(\vec{s}_1) = \vec{k}(\vec{s}_2) = \vec{y}$ or not. 
This is why the existence of $\mathcal{F}$ is not sufficient by itself.

It is important to note that both $\vec{k}$ and $\vec{\mathcal{F}}$ are unknown in our problem setting, so that the theorem only guarantees the possibility of finding the right $\vec{\psi}$ but does not provide a constructive way of finding it.
Therefore, when learning is not realizable it is generally impossible to understand whether the problem is related to $\vec{\mathcal{F}}$, to $\vec{\psi}$, or even to the both of them.

Notably, as we will discuss later, this problem can be bypassed by considering the synchronization function $\vec{\phi}$ as a surrogate for $\vec{\mathcal{F}}$. As $\vec{\phi}$ is only related to the dynamical evolution of the reservoir (listening phase), we can discuss its existence and properties disregarding the readout.

This shows the importance of $\vec{\phi}$ in the context of \ac{RC}: it can be used to asses the quality of the representation of the unknown source system that the reservoir has encoded in its state. 
This allows one to disentangle the problem of embedding the input (which is done in an unsupervised way during the listeining phase) from the the problem of finding the best readout to predict the target (which is a supervised problem, faced in the fitting phase).
This fact is of particular interest as most of the hyperparameters that are usually optimized (e.g., the spectral radius of the connectivity matrix, its sparsity, the input scaling, the activation function) affect the listening phase only and, therefore, the synchronization. Hence, their analysis and optimization  can be performed disregarding the fitting procedure.

Finally, we point out that Theorem~\ref{thm:sufficient_realizability} formally proves that -- as suggested in other works \cite{lu2018attractor, lu2020invertible} -- the existence of an invertible synchronization function is sufficient for the \ac{RC} paradigm to work (provided that the readout is able to correctly approximate the target). We proved that this condition applies not only in the generative frameworks (i.e., when $\vec{y}(t) = \vec{s}(t+1)$) which is the one studied in \cite{lu2018attractor, lu2020invertible}, but to any generic target  $\vec{y}(t) = \vec{k}(\vec{s}(t))$.

\subsection{Error on the whole attractor}

Since the readout $\vec{\psi}$ is generated after the listening phase, we have no guarantees that, in general, it will continue to correctly reproduce the target also in the predicting phase.
More in detail, after observing a series of measurements $\vec{u}(t)$ and targets $\vec{y}(t)$ coming from an \emph{unknown} trajectory of the source system $\vec{s}(t)$, we want to learn a readout $\vec{\psi}$ which is able to predict the targets even for future times.

Since we have assumed that the source system \eqref{eqn:source_system} has a unique attractor $\mathcal{A}$, this goal can be achieved by learning a readout valid for all the $\vec{y} = \vec{k}(\vec{s})$, for $\vec{s} \in \mathcal{A}$.
In the machine learning parlance, this can be described as follows: a single trajectory plays the role of a sample, while the attractor plays the role of the data-generating process.
This becomes possible by assuming the attractor $\mathcal{A}_s$ to be ergodic \cite{birkhoff1931proof}. 
In fact, the existence of an ergodic attractor guarantees that a sufficiently long trajectory will be a ``good sampling'' of the whole attractor (see \cite{hart2020echo} for a discussion in the generative framework).
Moreover, as all trajectories starting from the basin of attraction $\mathcal{B}_s$ will approach $\mathcal{A}_s$, this procedure allows to learn a prediction model suitable for a full set of trajectories by observing only one.

To do so, let us define the loss function: 
\begin{equation} \label{eqn:RMSE}
\mathcal{L}(\vec{y}(t), \hat{\vec{y}}(t)) =  \lVert \vec{y}(t) - \hat{\vec{y}}(t) \rVert_2
\end{equation}
where $\lVert \cdot \rVert_2$ is the $L_2$-norm. We refer to \eqref{eqn:RMSE} as \ac{RMSE}.%
\footnote{Note that different choices can be made for $\mathcal{L}$ and the results do not depend on its particular form. We use the \ac{RMSE} because it is the one we use in the esperimental section.}
The learning realizability trivially implies that there exists a readout for which:
\begin{equation}
    \frac{1}{T} \sum_{t=t_s}^{T} \mathcal{L}(\vec{y}(t), \hat{\vec{y}}(t)) = 0
\end{equation}
since $\mathcal{L}(\vec{y}(t), \hat{\vec{y}}(t)) = 0, \forall t$.
Note that time starts at $t= t_s$ because we want to remove transient effects (as our discussion is valid on the attractor only).

By expanding $\vec{y}$ and $\hat{\vec{y}}$, we get:
\begin{equation}
    \frac{1}{T} \sum_{t=t_s}^{T}  \mathcal{L}(\vec{k}(\vec{s}(t)), \vec{\psi}(\vec{r}(t))) = 0
\end{equation}

The existence of a function $\vec{\mathcal{F}}$ (Thm. \ref{thm:sufficient_realizability}) allows us to write  $\vec{r}(t) = \vec{\mathcal{F}}(\vec{s}(t))$, so that our loss becomes

\begin{equation}
\mathcal{L}(\vec{k}(\vec{s}(t)), \vec{\psi}(\vec{r}(t))) = \mathcal{L}(\vec{k}(\vec{s}(t)), \vec{\psi}(\vec{\mathcal{F}}(\vec{s}(t)))) = 
\mathcal{L}(\vec{s}(t))
\end{equation}
where the last equality stresses the fact that $\mathcal{L}$ is a function of $\vec{s}(t)$ only (with an abuse of notation on the function $\mathcal{L}$).
Taking the limit for $T \to \infty$, we can now exploit the ergodicity of $\mathcal{A}_s$ and obtain:
\begin{equation}
\label{eq:error_whole_attractor_As}
    0 =\lim_{T \to \infty}\frac{1}{T} \sum_{t=t_s}^{T} \mathcal{L}(\vec{y}(t), \hat{\vec{y}}(t)) 
    = \underbrace{\lim_{T \to \infty}\frac{1}{T}\sum_{t=t_s}^{T} \mathcal{L}(\vec{s}(t)) = \langle \mathcal{L}(\vec{s}) \rangle_{\mathcal{A}_s}}_{\mathrm{ergodicity}}
\end{equation}
In \eqref{eq:error_whole_attractor_As}, $\langle \mathcal{L}(\vec{s}) \rangle_{\mathcal{A}_s}$ denotes the average loss by sampling trajectories over the whole attractor $\mathcal{A}_s$.
This means that, if learning is realizable for a single trajectory, then it will be realizable on the whole attractor of the source system.

Note that the crucial part of this approach is the dependence on $\vec{s}$ only, because only the source system attractor $\mathcal{A}_s$ is assumed to be ergodic.

\subsection{Synchronization function}

One would like to relax the definition of realizable learning (see Def.~\ref{def:realizability}): in fact, in realistic situations the error is not exactly zero. This is formalized by assuming that $\mathcal{L}(\vec{y}(t), \hat{\vec{y}}(t)) = \epsilon_t \ge 0, \forall t$, so that:
\begin{equation}
    \mathcal{E}_{T} = \frac{1}{T} \sum_{t=t_s}^{T} \mathcal{L}(\vec{y}(t), \hat{\vec{y}}(t))
\end{equation}

As proved in the previous section, if a mapping $\vec{\mathcal{F}}$ does not exist, then learning cannot be realizable according to Def.~\ref{def:realizability}.
But assuming \ac{GS} to hold, we can make use of the synchronization function $\vec{\phi}$ and write:
\begin{equation}
\begin{aligned}
\frac{1}{T} \sum_{t=t_s}^{T} \mathcal{L}(\vec{y}(t), \hat{\vec{y}}(t)) 
&=\frac{1}{T} \sum_{t=t_s}^{T} \mathcal{L}(\vec{k}(\vec{s}(t)), \vec{\psi}(\vec{\phi}(\vec{s}(t)))) \\
&= \frac{1}{T} \sum_{t=t_s}^{T} \mathcal{L}(\vec{s}(t))
\end{aligned}
\end{equation}
where, again, $\mathcal{L}$ depends only on $\vec{s}(t)$.
In order to make use of the ergodicity of the source system, we need to be sure that the above limit exists. 
An easy way for guaranteeing this consists of requiring the error to be bounded, i.e., to have $\mathcal{L}(\vec{y}(t), \hat{\vec{y}}(t)) = \epsilon_t < C$, where $C \ge 0$  is a constant.
So, when $\lim_{T \to \infty} E_{T}$ exists and is finite, one can write:
\begin{equation}
    \mathcal{E}=
    \lim_{T \to \infty} \mathcal{E}_{T}
    = \lim_{T \to \infty} \frac{1}{T} \sum_{t=t_s}^{T} \mathcal{L}(\vec{s}(t))
    = \langle \mathcal{L}(\vec{s}(t)) \rangle_{\mathcal{A}_s}
\end{equation}

The existence of the synchronization function guarantees that the error for a single trajectory is the same as the error in the whole attractor of the source system.
This means that, by assuming \ac{GS}, we can have some guarantees on the performance of our model even when learning is not realizable, and this is due to the fact that $\mathcal{L}$ depends only on the source system states $\vec{s}$ when assuming \ac{GS}. 
We emphasize that this argument applies not only to future time of the the same trajectory, but also to any trajectory of the source system \eqref{eqn:source_system} which starts from $\mathcal{B}_s$.
Practically speaking, this means that if we have a trajectory of the source system starting at $\vec{s}'(0) \in \mathcal{B}_s$, the readout that was previously trained will still work, because this new trajectory will still approach the same attractor $\mathcal{A}_s$ and the reservoir will approach the synchronization manifold.
Note that if the \ac{GS} is granted but it is not unique (i.e, we do not have the \ac{ESP}), the reservoir needs to be initialized with the same initial condition, otherwise it may converge to a different synchronization manifold, which would require a different readout; which always exists as long as the new synchronization function is invertible. The \ac{ESP} ensures that the readout will be the same, as the synchronization function will be unique.
In fact, the definition of the synchronization function in Def.~\ref{def:GS} has other implications for the training mechanisms.
Making use of its smoothness along with the attractivity of the synchronization manifold, one can account not only for the error in the approximation, but also for the observational noise of the source system (see Appendix~\ref{sec:noise} for details).

Finally, we stress that the existence of \ac{GS} is a property which only involves the source system \eqref{eqn:source_system} and its coupling with the reservoir \eqref{eqn:listening} by means of the measurements \eqref{eqn:measurement}, disregarding the particular task at hand.
In fact, in the discussion above, we showed how using the synchronization function $\vec{\phi}$ one can, in some sense, decouple the learning task and separate the problem of finding a suitable readout from the problem of granting the existence of a mapping from the source system states, $\vec{s}$, to the reservoir states, $\vec{r}$.

\section{Experimental results}
\label{sec:experiments}

\subsection{The mutual false nearest neighbors}

Identifying \ac{GS} is hard due to the fact that the synchronization function \eqref{eqn:GS_condition} in unknown and may take any form. For this reason, in \cite{rulkov1995generalized} a method to empirically assess the occurrence of \ac{GS} from data was proposed under the name of \ac{MFNN}.
It is based on the fact that, under reasonable smoothness conditions for $\vec{\phi}$, \ref{eqn:GS_condition} implies that two states that are close in state-space of the response system correspond to two close states in the state-space of the driving system.
So, we are looking for a geometric connection between the two systems which preserves the neighbor-structure in state space.

Let us assume that we sample trajectories from a dynamical system at a fixed sampling rate, resulting in a series of discrete times $\{t_n\}$.
The resulting measurements for the two systems will be $\{\vec{x}_n\}$ and $\{\vec{y}_n\}$, for the drive and the response respectively, where we used the notation $\vec{x}_n := \vec{x}(t_n)$ and $\vec{y}_n := \vec{y}(t_n)$.
For each point $\vec{x}_n$ of the driving system, we seek the closest point from its neighbors, which we will call time index $n_\text{NND}$. 
Then, due to \eqref{eqn:GS_condition}, the point $\vec{y}_n = \vec{\phi}(\vec{x}_n)$ will be close to $\vec{y}_{n_\text{NND}}$.
If the distances between these pairs of points in state-space of both the drive and response systems are small, one can write:
\begin{equation}\label{eqn:NND_approx}
    \vec{y}_n - \vec{y}_{n_\text{NND}} 
    = \vec{\phi}(\vec{x}_n) - \vec{\phi}(\vec{x}_{n_\text{NND}})
    \approx \matr{D}\vec{\phi}(\vec{x}_n) (\vec{x}_n - \vec{x}_{n_\text{NND}})
\end{equation}
where $\matr{D}\vec{\phi}(\vec{x}_n)$ is the Jacobian of $\vec{\phi}$ evaluated at $\vec{x}_n$.

Now, we do a similar operation but in the response state space.
We look for the closer point to $\vec{y}_n$ and we index it with $n_\text{NNR}$. 
Again, due to \eqref{eqn:GS_condition}, it holds:
\begin{equation}\label{eqn:NNR_approx}
    \vec{y}_n - \vec{y}_{n_\text{NNR}} 
    = \vec{\phi}(\vec{x}_n) - \vec{\phi}(\vec{x}_{n_\text{NNR}})
    \approx \matr{D}\vec{\phi}(\vec{x}_n) (\vec{x}_n - \vec{x}_{n_\text{NNR}})
\end{equation}

So, due to \eqref{eqn:NND_approx} and \eqref{eqn:NNR_approx} we have two different ways of evaluating $\matr{D}\vec{\phi}(\vec{x}_n)$.
This leads us to the definition of the \ac{MFNN} as the following ratio:
\begin{equation}\label{eqn:MFNN}
    \text{MFNN}(n) := 
    \frac{\lVert  \vec{y}_n - \vec{y}_{n_\text{NND}} \rVert}
        {\lVert  \vec{x}_n - \vec{x}_{n_\text{NND}} \rVert}
    \frac{\lVert  \vec{x}_n - \vec{x}_{n_\text{NNR}} \rVert}
        {\lVert  \vec{y}_n - \vec{y}_{n_\text{NNR}} \rVert}
\end{equation}

If the two systems are synchronized in a general sense, then $\text{MFNN}(n) \approx 1$.
If the synchronization relation does not hold, then \eqref{eqn:MFNN} should instead be of the order of (size of the attractor squared)/(distance between nearest neighbors squared) which is, in general, a large number.

Note that in this work we use the full knowledge of the source system to measure the \ac{GS} by means of \ac{MFNN}.
Generally, one would not have such a knowledge: anyway the \ac{MFNN} can be used also in this case, as showed in the paper where it was proposed \cite{rulkov1995generalized}, making use of the embedding theorem.
For simplicity, we do not deal with this more complex case, since it would not be relevant for the discussion.

Another possible way of assessing \ac{GS} is to verify the complete synchronization (see Appendix~\ref{sec:complet_synch}) between multiple copies of the reservoir. While such an approach might be hard to follow when considering physical systems, it poses no problem and can be straightforwardly implemented in the context we are interested in (\cite{platt2021forecasting}).

\subsection{Reservoir computing networks}
\label{sec:rcn}

For simplicity, but without loss of generality, we will only deal with one-dimensional inputs $u(t)\in\mathbb{R}$.
We use an \ac{RCN} where the explicit form of the reservoir equation \eqref{eqn:listening} reads:
\begin{equation} \label{eqn:RCN_state}
    \vec{r}(t + \tau) = \tanh \left(\matr{W} \vec{r}(t) +  \vec{w} u(t+\tau)  + \vec{b} \right)
\end{equation}
$\matr{W} \in \mathbb{R}^{d_r \times d_r}$ is the \emph{connectivity matrix}, which is an Erdos-Renyi matrix with average degree $6$; $d_r$ indicates the dimension of the reservoir.
The non-null elements are drawn from a uniform distribution taking values in the interval $(-1, 1)$.
$\matr{W}$ is re-scaled so that its \emph{\ac{SR}} equals the user-defined hyper-parameter $\rho>0$.
{We emphasize that having a \ac{SR} smaller than $1$ is not a sufficient nor necessary condition for the \ac{ESP} to hold~\cite{yildiz2012re}.}
The elements $\vec{w} \in \mathbb{R}^{d_r}$ of the \emph{input vector} are drawn from a uniform distribution taking values in $(-\omega, \omega)$; we refer to $\omega$ as the \emph{input scaling} hyper-parameter.
$\vec{b} = [b,b, \dots, b]$ is a constant bias term, which is useful to control the non-linearity of the system and to break the symmetry with respect to the origin \cite{lu2017reservoir}. 
Here, $\tanh$ stands for the hyperbolic-tangent function applied element-wise.

We use a linear readout, so that the predicted output is given by:
\begin{equation} \label{eqn:RCN_readout}
    \hat{\vec{y}}(t) = \vec{\psi}(\vec{r}(t)) \equiv \matr{W}_{\text{out}} \vec{r}(t)
\end{equation}
where $\matr{W}_{\text{out}}$ is a $d_y \times d_r$ matrix, called readout matrix; $d_y$ is the output dimension.
We train the readout using ridge-regression with regularization parameter $\lambda$, but more sophisticated, offline optimization procedures can be designed as well \cite{gallicchio2017deep, shi2007support, lokse2017training}.

\subsection{Reservoir observer}\label{subsec:Reservoir_Observer}

We test the hypothesis that learning in \ac{RC} can happen only when the network is synchronized with the source system \eqref{eqn:source_system}.
To do so, we adopt the framework named \emph{reservoir observer} \cite{lu2017reservoir}, which consists of setting $\vec{h}(\vec{s}) = s_1 = u$ and $\vec{y} = \vec{s}$. This means that the network is trained to reconstruct the full state of the source system by seeing only one component of it.
Note that $\vec{k}$ in \eqref{eqn:target} is the identity for this task, and \eqref{eqn:readout_true} reduces to finding the right inverse of the synchronization function \eqref{eqn:GS_condition}.
Practically, when using a linear readout (as in fact we do here), one implicitly assumes that $\vec{k} \circ \vec{\phi}^{\dagger}_{}(\vec{r})$ in \eqref{eqn:readout_true} can be expressed in linear form 
\begin{equation}
\vec{\phi}^{\dagger}_{} (\vec{r}) = \matr{W}_\text{out} \vec{r}
\end{equation}
implying that
\begin{equation}
\vec{\phi}_{} (\vec{s}) = \matr{W}_\text{out}^{*} \vec{s}
\end{equation}
where $\matr{W}_\text{out}^{*}$ is the left pseudo-inverse of the readout matrix.

\subsection{Results}\label{sec:results}

\begin{figure}[ht!]
    \centering
    \includegraphics[width =.8 \textwidth]{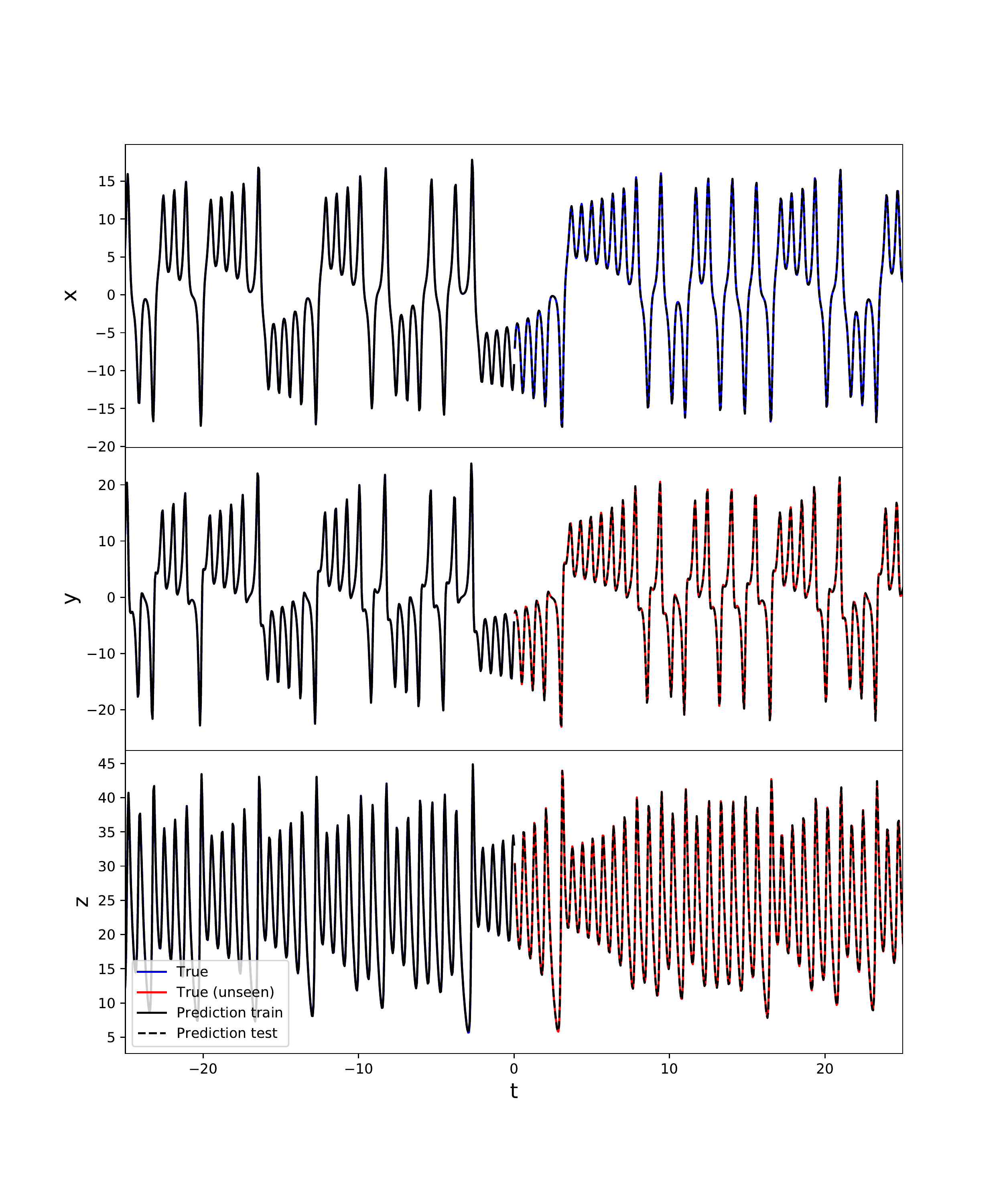}
    \caption{
    An example of the observer task using  the Lorenz system. 
    The top panel show the measurement $\vec{u}$ (blue), which is always available.
    The middle and the bottom panels represents the targets $\vec{y}$: in the training phase they are available (blue) while in the predicting phase (red) they cannot be accessed anymore.
    The predicted targets $\hat{\vec{y}}$ (black dashed lines) are generated by means of the \ac{RCN} described in Sec.~\ref{sec:rcn}.}
    \label{fig:observer_example}
\end{figure}

As for the source system \eqref{eqn:source_system}, we use the Lorenz model (see Appendix \ref{sec:lorenz} for details). 
In the listening phase, we use $t$ in the interval of $T_\text{train} = (-100,0)$.
We discarded the first $1/10$ of the data points for training, to account for transient effects in the synchronization process.
The prediction phase was carried out for values of $t$ in $T_\text{test}=(0,80)$. 
The integration step was always set to $\tau = 0.05$.
An example of this task is provided in Fig.~\ref{fig:observer_example}.

For each hyper-parameter configuration, we repeated the experiment $10$ times, generating a different realization of the source system (i.e., starting from distinct random initial conditions) and a different realization of the \ac{RCN} \eqref{eqn:RCN_state}. 
For each run, the \ac{MFNN} between the driving system state $\vec{s}(t)$ and the reservoir $\vec{r}(t)$ was computed.
As a performance measure for the prediction accuracy, we used the \ac{RMSE} computed on the $y$ and the $z$ coordinate of the Lorenz system (since $x$ is used as input).
Following~\cite{rulkov1995generalized}, we plot the inverse of the \ac{MFNN} so that the higher the value, the more synchronized the systems are. 
Accordingly, we plot the inverse of the \ac{RMSE}, which can be interpreted as a form of accuracy.
Unless differently stated, all hyperparameters are the ones reported in Tab.~\ref{tab:default_params}.
All the plots refer to the \emph{predicting phase}.
\begin{table}[h]
    \centering
    \caption{Default hyper--parameters used in all experiments, unless differently stated.}
    \begin{tabular}{LL|LL}
        \rho   & 1   & T_\text{train} & (-100,0) \\
        \omega & 0.1 & T_\text{test}           & (0,80)   \\
        d_r   & 300 & \tau                    & 0.05     \\
        b      & 1   & \lambda                 & 10^{-6}  \\ 
    \end{tabular}
    \label{tab:default_params}
\end{table}

\begin{figure}[h]
    \centering
    \includegraphics[width = \textwidth]{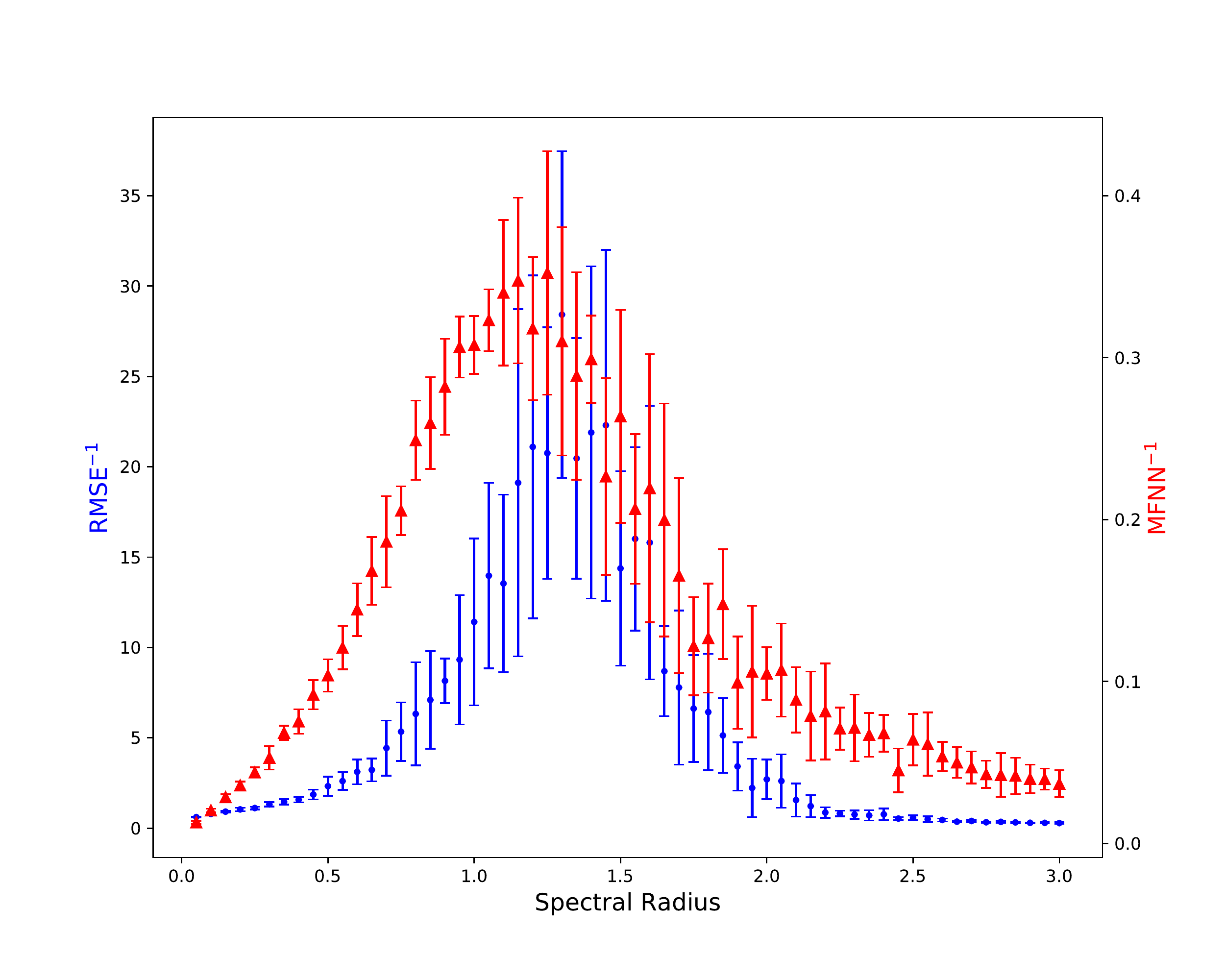}
    \caption{Results for the reservoir observer when varying the \ac{SR} of the connectivity matrix, when using the Lorenz system as a source.
    Blue dots account for the \ac{RMSE} (left axis) while red triangles accounts for \ac{MFNN} (right axis). }
    \label{fig:observer_lorenz}
\end{figure}

In Fig.~\ref{fig:observer_lorenz}, we show the \ac{RMSE} and the \ac{MFNN} index when the \ac{SR} of the reservoir connectivity matrix varies in a suitable range. 
For smaller values of \ac{SR} the reservoir dynamics are really simple and close to linear (since $\tanh(x) \approx x$ when  $x$ is small), so that the network it is not able to correctly represent the Lorenz attractor in its state. 
We see that the synchronization is weak and the error is large. 
As the \ac{SR} approaches $1$ we see that the reservoir tends to become more synchronized with the source system state and this reflects in a smaller error. 
When the \ac{SR} started growing, the reservoirs becomes more and more unstable and it gradually de-synchronizes with the source system, such that the reconstruction of the coordinates becomes less precise.

\begin{figure}[h]
    \centering
    \includegraphics[width = \textwidth]{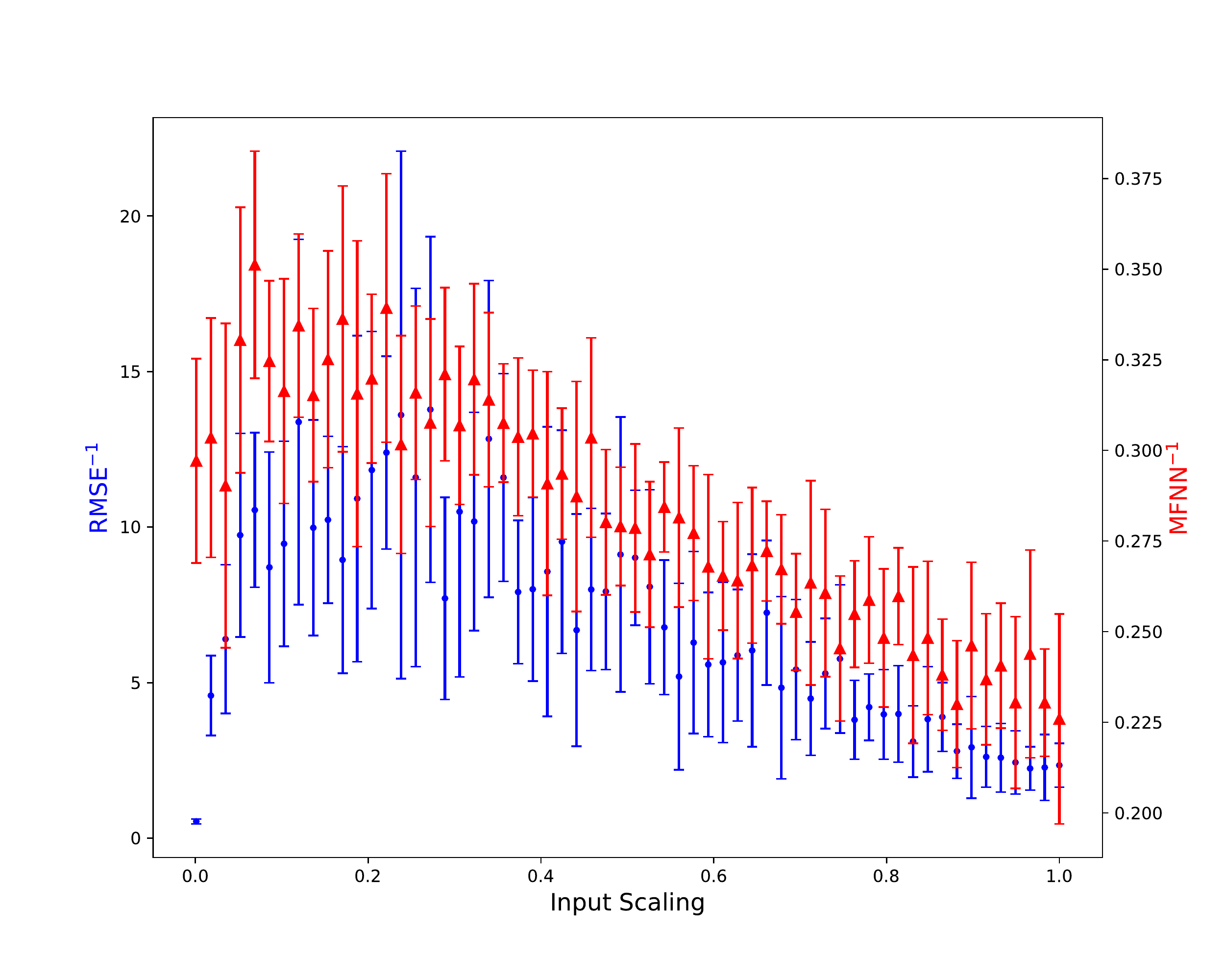}
    \caption{Results for the reservoir observer  on Lorenz system when varying the input scaling $\omega$.
    when using the Lorenz system as a source.}
    \label{fig:Lorenz_omega}
\end{figure}

In Fig.~\ref{fig:Lorenz_omega} we repeated the experiment, but this time varying the input scaling $\omega$ and holding $\rho$ fixed to its default value.
We see that the two quantities still correlates, with both the accuracy and the synchronization decreasing as the input scaling grows.

\begin{figure}[h]
    \centering
    \includegraphics[width = \textwidth]{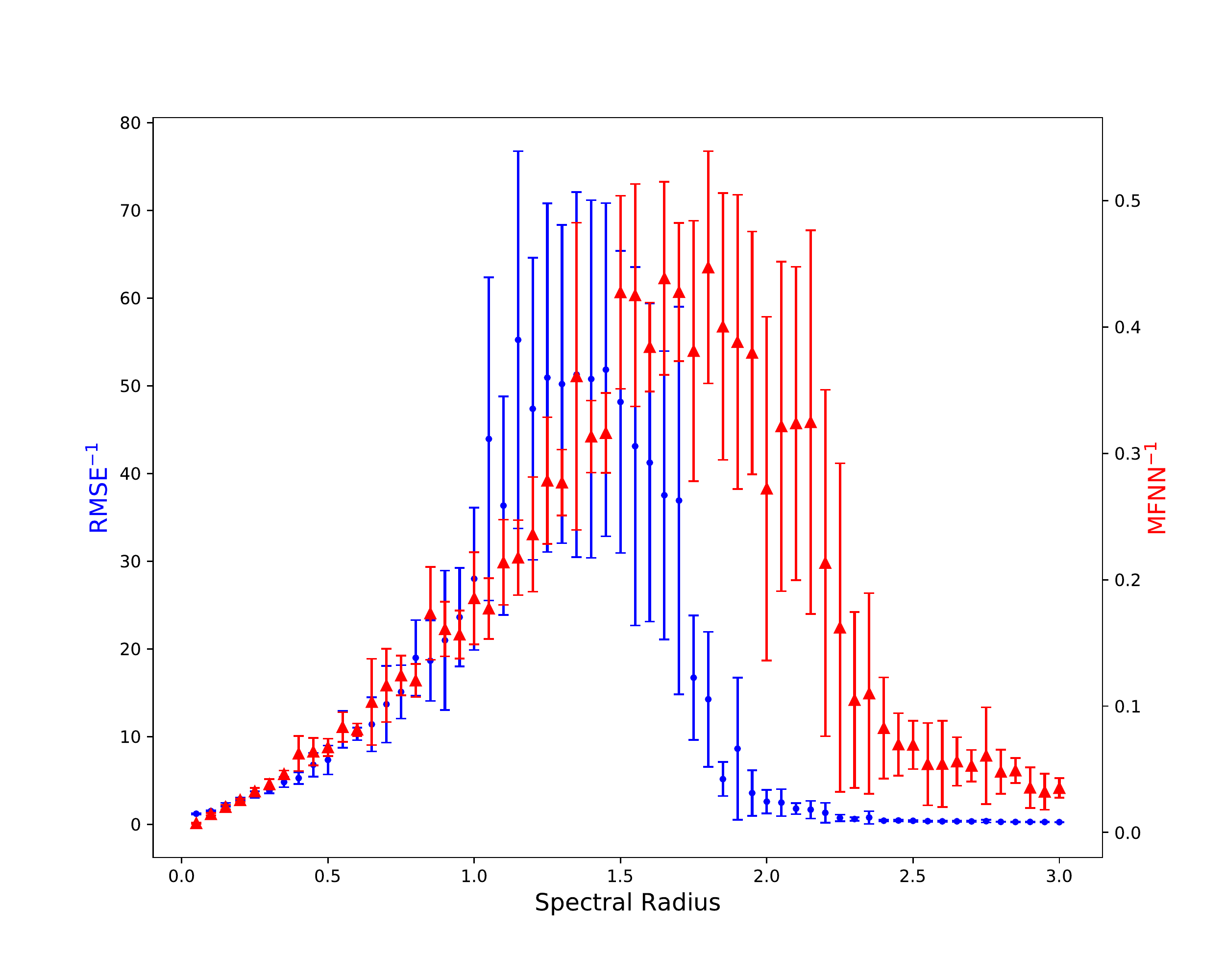}
    \caption{Results for the reservoir observer when varying the \ac{SR}, when using the R{\"o}ssler system as a source.
    Blue dots account for the \ac{RMSE} (left axis) while red triangles accounts for \ac{MFNN} (right axis).}
    \label{fig:observer_rossler}
\end{figure}

To assess the generality of our findings, we performed additional simulations by changing the source system \eqref{eqn:source_system}.
To this end, we consider now the R{\"o}ssler system as a source system (see Appendix \ref{sec:roessler} for details).
Since the dynamics of the R{\"o}ssler system are slower then the Lorenz ones, we set the integration step to $\tau = 0.5$, $T_\text{train}= (-200,0)$ and $T_\text{text}= (0,160)$.
The remaining hyper-parameters are set as shown in Tab.~\ref{tab:default_params}.
Again, we use the $x$-coordinate as input and the tasks consists of learning how to reproduce $y$ and $z$.
The results are displayed in Fig.~\ref{fig:observer_rossler} and look similar to the one obtained for the Lorenz system (Fig.~\ref{fig:observer_lorenz}).

To show that \ac{GS} plays an important role not only in the observer task, we also test our framework in a forecasting scenario.
To this end, we use the $x$-coordinate of the Lorenz system as the input ($u(t) := x(t)$, but this time the target $\vec{y}$ was chosen to be $y(t) := u(t+ 5\tau)$.
This means that the \ac{RCN} is required to correctly approximate the function $\vec{g}^5(\vec{s})$, which is highly nonlinear.
The \ac{RMSE} is computed between $y(t)$ and the network output $\hat{y}(t)$.
Notably, the \ac{MFNN} here is almost identical to the one in Fig.~\ref{fig:observer_lorenz}: both experiments use the same hyperparameters, the same source system and construct $\vec{u}$ in the same way, so that the only difference (up to the particular realization) is the task they are trained to solve, which affects the readout and not on the dynamics.
As in the other cases, we notice that the \ac{MFNN} and the \ac{RMSE} display a similar behavior.
\begin{figure}[h]
    \centering
    \includegraphics[width = \textwidth]{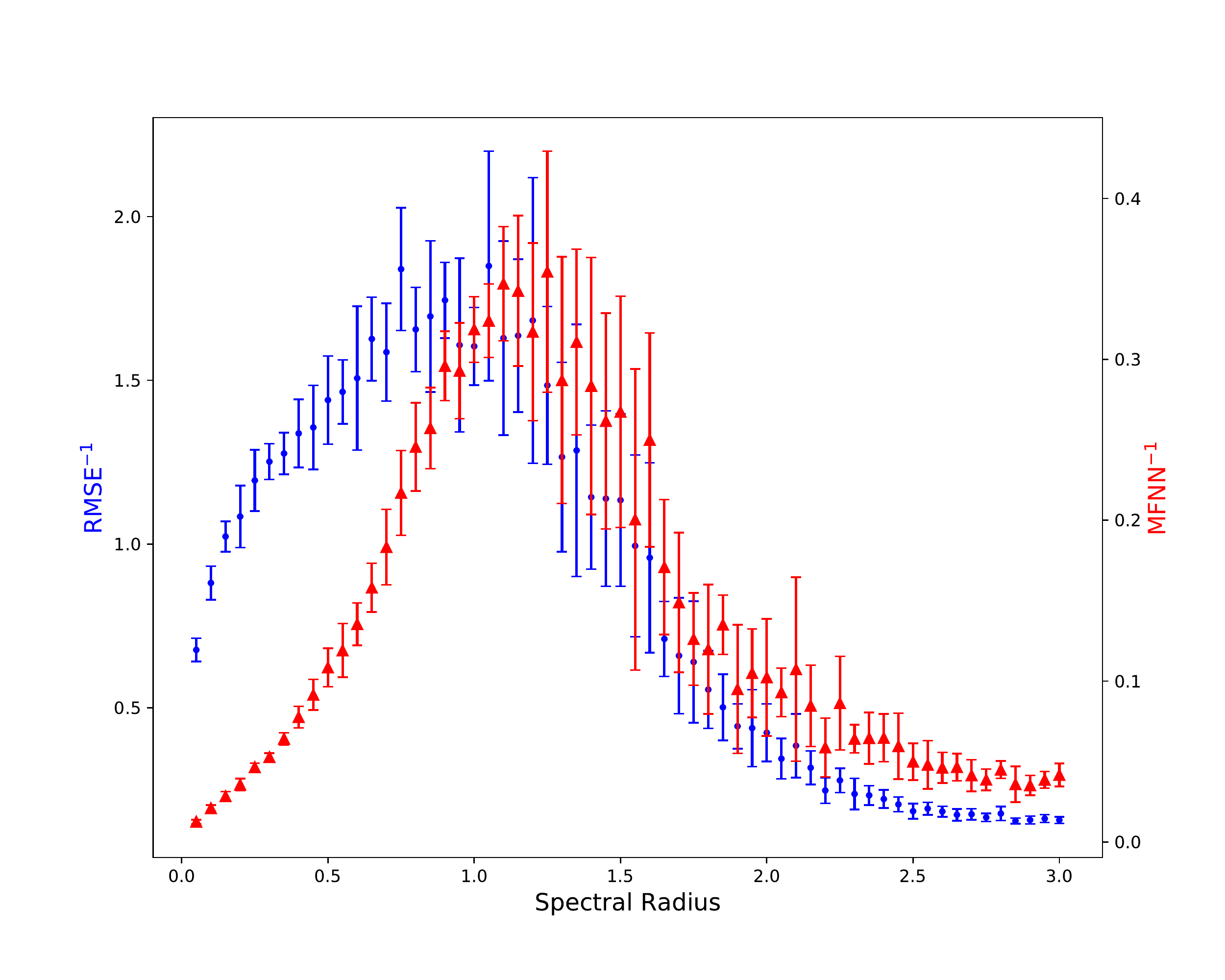}
    \caption{Results for the forecasting task on Lorenz system when varying the \ac{SR}.}
    \label{fig:Prediction}
\end{figure}

These results confirm that the \ac{GS} can be exploited to assess the quality of the source system representation encoded in the reservoir states: in order to correctly solve the task at hand, the reservoir and the the source system should be synchronized.

\section{Conclusions}\label{sec:conclusions}

In this work, we laid down the groundwork for establishing and analyzing working principles of \ac{RC} within the theoretical framework of synchronization between dynamical systems.
First, we made systematic the equivalence between the \ac{ESP} and \ac{GS}.
Then, we showed that the presence of a synchronization function consents to formally consider the reservoir states as an unsupervised, high-dimensional representation of an unknown source system that generates the observed data.
We showed that the realizability of learning, defined as the possibility of perfectly solving the task, crucially depends of the existence of a function connecting the reservoir states with the source system states:
the presence of \ac{GS} implies the existence of a synchronization function playing an analogous role, which is found in an unsupervised way in \ac{RC}.
This formally proves that it is possible to solve the task at hand by firstly creating an unsupervised representation of the source system (listening phase) and then using a suitable readout to correctly represent the target (fitting phase), thus justifying the \ac{RC} training principle in a formal way.
Moreover, the presence of such a synchronization function allows one to make use of the ergodicity of the source system to grant some results on the generalization error for a given task.
Finally, we made use of an index (\ac{MFNN}) to quantify the degree of synchronization and experimentally validate our claims. Results show that the more the reservoir is synchronized with the source, the better the system approximates the target, hence stressing that synchronization is paramount and plays a fundamental role within the \ac{RC} framework.

\section*{Declaration of Competing Interest}

The authors declare that they have no known competing financial interests or personal relationships that could have appeared to influence the work reported in this paper.

\section*{Acknowledgments}

LL gratefully acknowledges partial support of the Canada Research Chairs program.  PV would like to thank Andrea Ceni and Daniele Zambon for the insightful discussions about some mathematical details of this work.

\bibliographystyle{elsarticle-num-names}
\bibliography{biblio.bib}

\begin{thebibliography}{62}
\expandafter\ifx\csname natexlab\endcsname\relax\def\natexlab#1{#1}\fi
\providecommand{\url}[1]{\texttt{#1}}
\providecommand{\href}[2]{#2}
\providecommand{\path}[1]{#1}
\providecommand{\DOIprefix}{doi:}
\providecommand{\ArXivprefix}{arXiv:}
\providecommand{\URLprefix}{URL: }
\providecommand{\Pubmedprefix}{pmid:}
\providecommand{\doi}[1]{\href{http://dx.doi.org/#1}{\path{#1}}}
\providecommand{\Pubmed}[1]{\href{pmid:#1}{\path{#1}}}
\providecommand{\bibinfo}[2]{#2}
\ifx\xfnm\relax \def\xfnm[#1]{\unskip,\space#1}\fi
\bibitem[{Liu and Theodorou(2019)}]{liu2019deep}
\bibinfo{author}{G.-H. Liu}, \bibinfo{author}{E.~A. Theodorou},
\newblock \bibinfo{title}{Deep learning theory review: An optimal control and
  dynamical systems perspective},
\newblock \bibinfo{journal}{arXiv preprint arXiv:1908.10920}
  (\bibinfo{year}{2019}).
\bibitem[{Bianchi et~al.(2018)Bianchi, Livi, and
  Alippi}]{bianchi2016investigating}
\bibinfo{author}{F.~M. Bianchi}, \bibinfo{author}{L.~Livi},
  \bibinfo{author}{C.~Alippi},
\newblock \bibinfo{title}{Investigating echo state networks dynamics by means
  of recurrence analysis},
\newblock \bibinfo{journal}{IEEE Transactions on Neural Networks and Learning
  Systems} \bibinfo{volume}{29} (\bibinfo{year}{2018})
  \bibinfo{pages}{427--439}. \DOIprefix\doi{10.1109/TNNLS.2016.2630802}.
\bibitem[{Sussillo and Abbott(2009)}]{sussillo2009generating}
\bibinfo{author}{D.~Sussillo}, \bibinfo{author}{L.~F. Abbott},
\newblock \bibinfo{title}{Generating coherent patterns of activity from chaotic
  neural networks},
\newblock \bibinfo{journal}{Neuron} \bibinfo{volume}{63} (\bibinfo{year}{2009})
  \bibinfo{pages}{544--557}.
\bibitem[{Bengio et~al.(1994)Bengio, Simard, and Frasconi}]{bengio1994learning}
\bibinfo{author}{Y.~Bengio}, \bibinfo{author}{P.~Simard},
  \bibinfo{author}{P.~Frasconi},
\newblock \bibinfo{title}{Learning long-term dependencies with gradient descent
  is difficult},
\newblock \bibinfo{journal}{IEEE Transactions on Neural Networks}
  \bibinfo{volume}{5} (\bibinfo{year}{1994}) \bibinfo{pages}{157--166}.
\bibitem[{Bouvrie and Hamzi(2017)}]{bouvrie2017kernel}
\bibinfo{author}{J.~Bouvrie}, \bibinfo{author}{B.~Hamzi},
\newblock \bibinfo{title}{Kernel methods for the approximation of nonlinear
  systems},
\newblock \bibinfo{journal}{SIAM Journal on Control and Optimization}
  \bibinfo{volume}{55} (\bibinfo{year}{2017}) \bibinfo{pages}{2460--2492}.
  \DOIprefix\doi{10.1137/14096815X}.
\bibitem[{Qi and Majda(2020)}]{qi2020using}
\bibinfo{author}{D.~Qi}, \bibinfo{author}{A.~J. Majda},
\newblock \bibinfo{title}{Using machine learning to predict extreme events in
  complex systems},
\newblock \bibinfo{journal}{Proceedings of the National Academy of Sciences}
  \bibinfo{volume}{117} (\bibinfo{year}{2020}) \bibinfo{pages}{52--59}.
\bibitem[{Gilpin(2020)}]{gilpin2020deep}
\bibinfo{author}{W.~Gilpin},
\newblock \bibinfo{title}{Deep learning of dynamical attractors from time
  series measurements},
\newblock \bibinfo{journal}{arXiv preprint arXiv:2002.05909}
  (\bibinfo{year}{2020}).
\bibitem[{Tu et~al.(2013)Tu, Rowley, Luchtenburg, Brunton, and
  Kutz}]{tu2013dynamic}
\bibinfo{author}{J.~H. Tu}, \bibinfo{author}{C.~W. Rowley},
  \bibinfo{author}{D.~M. Luchtenburg}, \bibinfo{author}{S.~L. Brunton},
  \bibinfo{author}{J.~N. Kutz},
\newblock \bibinfo{title}{On dynamic mode decomposition: Theory and
  applications},
\newblock \bibinfo{journal}{arXiv preprint arXiv:1312.0041}
  (\bibinfo{year}{2013}).
\bibitem[{Berry et~al.(2020)Berry, Giannakis, and Harlim}]{berry2020bridging}
\bibinfo{author}{T.~Berry}, \bibinfo{author}{D.~Giannakis},
  \bibinfo{author}{J.~Harlim},
\newblock \bibinfo{title}{Bridging data science and dynamical systems theory},
\newblock \bibinfo{journal}{arXiv preprint arXiv:2002.07928}
  (\bibinfo{year}{2020}).
\bibitem[{Verstraeten et~al.(2007)Verstraeten, Schrauwen, d'Haene, and
  Stroobandt}]{verstraeten2007experimental}
\bibinfo{author}{D.~Verstraeten}, \bibinfo{author}{B.~Schrauwen},
  \bibinfo{author}{M.~d'Haene}, \bibinfo{author}{D.~Stroobandt},
\newblock \bibinfo{title}{An experimental unification of reservoir computing
  methods},
\newblock \bibinfo{journal}{Neural Networks} \bibinfo{volume}{20}
  (\bibinfo{year}{2007}) \bibinfo{pages}{391--403}.
\bibitem[{Lu et~al.(2017)Lu, Pathak, Hunt, Girvan, Brockett, and
  Ott}]{lu2017reservoir}
\bibinfo{author}{Z.~Lu}, \bibinfo{author}{J.~Pathak},
  \bibinfo{author}{B.~Hunt}, \bibinfo{author}{M.~Girvan},
  \bibinfo{author}{R.~Brockett}, \bibinfo{author}{E.~Ott},
\newblock \bibinfo{title}{Reservoir observers: Model-free inference of
  unmeasured variables in chaotic systems},
\newblock \bibinfo{journal}{Chaos: An Interdisciplinary Journal of Nonlinear
  Science} \bibinfo{volume}{27} (\bibinfo{year}{2017}) \bibinfo{pages}{041102}.
\bibitem[{Lu et~al.(2018)Lu, Hunt, and Ott}]{lu2018attractor}
\bibinfo{author}{Z.~Lu}, \bibinfo{author}{B.~R. Hunt},
  \bibinfo{author}{E.~Ott},
\newblock \bibinfo{title}{Attractor reconstruction by machine learning},
\newblock \bibinfo{journal}{Chaos: An Interdisciplinary Journal of Nonlinear
  Science} \bibinfo{volume}{28} (\bibinfo{year}{2018}) \bibinfo{pages}{061104}.
\bibitem[{Chattopadhyay et~al.(2020)Chattopadhyay, Hassanzadeh, and
  Subramanian}]{chattopadhyay2020data}
\bibinfo{author}{A.~Chattopadhyay}, \bibinfo{author}{P.~Hassanzadeh},
  \bibinfo{author}{D.~Subramanian},
\newblock \bibinfo{title}{Data-driven predictions of a multiscale lorenz 96
  chaotic system using machine-learning methods: reservoir computing,
  artificial neural network, and long short-term memory network},
\newblock \bibinfo{journal}{Nonlinear Processes in Geophysics}
  \bibinfo{volume}{27} (\bibinfo{year}{2020}) \bibinfo{pages}{373--389}.
\bibitem[{Vlachas et~al.(2020)Vlachas, Pathak, Hunt, Sapsis, Girvan, Ott, and
  Koumoutsakos}]{vlachas2020backpropagation}
\bibinfo{author}{P.~R. Vlachas}, \bibinfo{author}{J.~Pathak},
  \bibinfo{author}{B.~R. Hunt}, \bibinfo{author}{T.~P. Sapsis},
  \bibinfo{author}{M.~Girvan}, \bibinfo{author}{E.~Ott},
  \bibinfo{author}{P.~Koumoutsakos},
\newblock \bibinfo{title}{Backpropagation algorithms and reservoir computing in
  recurrent neural networks for the forecasting of complex spatiotemporal
  dynamics},
\newblock \bibinfo{journal}{Neural Networks}  (\bibinfo{year}{2020}).
\bibitem[{Bompas et~al.(2020)Bompas, Georgeot, and
  Gu{\'e}ry-Odelin}]{bompas2020accuracy}
\bibinfo{author}{S.~Bompas}, \bibinfo{author}{B.~Georgeot},
  \bibinfo{author}{D.~Gu{\'e}ry-Odelin},
\newblock \bibinfo{title}{Accuracy of neural networks for the simulation of
  chaotic dynamics: precision of training data vs precision of the algorithm},
\newblock \bibinfo{journal}{arXiv preprint arXiv:2008.04222}
  (\bibinfo{year}{2020}).
\bibitem[{Jaeger(2001)}]{jaeger2001echo}
\bibinfo{author}{H.~Jaeger},
\newblock \bibinfo{title}{The “echo state” approach to analysing and
  training recurrent neural networks-with an erratum note},
\newblock \bibinfo{journal}{Bonn, Germany: German National Research Center for
  Information Technology GMD Technical Report} \bibinfo{volume}{148}
  (\bibinfo{year}{2001}) \bibinfo{pages}{13}.
\bibitem[{Maass et~al.(2002)Maass, Natschl{\"a}ger, and
  Markram}]{maass2002real}
\bibinfo{author}{W.~Maass}, \bibinfo{author}{T.~Natschl{\"a}ger},
  \bibinfo{author}{H.~Markram},
\newblock \bibinfo{title}{Real-time computing without stable states: A new
  framework for neural computation based on perturbations},
\newblock \bibinfo{journal}{Neural Computation} \bibinfo{volume}{14}
  (\bibinfo{year}{2002}) \bibinfo{pages}{2531--2560}.
\bibitem[{Ti{\v{n}}o and Dorffner(2001)}]{tino2001predicting}
\bibinfo{author}{P.~Ti{\v{n}}o}, \bibinfo{author}{G.~Dorffner},
\newblock \bibinfo{title}{Predicting the future of discrete sequences from
  fractal representations of the past},
\newblock \bibinfo{journal}{Machine Learning} \bibinfo{volume}{45}
  (\bibinfo{year}{2001}) \bibinfo{pages}{187--217}.
\bibitem[{Grigoryeva and Ortega(2018)}]{grigoryeva2018echo}
\bibinfo{author}{L.~Grigoryeva}, \bibinfo{author}{J.-P. Ortega},
\newblock \bibinfo{title}{Echo state networks are universal},
\newblock \bibinfo{journal}{Neural Networks} \bibinfo{volume}{108}
  (\bibinfo{year}{2018}) \bibinfo{pages}{495--508}.
\bibitem[{Hart et~al.(2020{\natexlab{a}})Hart, Hook, and
  Dawes}]{hart2019embedding}
\bibinfo{author}{A.~Hart}, \bibinfo{author}{J.~Hook},
  \bibinfo{author}{J.~Dawes},
\newblock \bibinfo{title}{Embedding and approximation theorems for echo state
  networks},
\newblock \bibinfo{journal}{Neural Networks} \bibinfo{volume}{128}
  (\bibinfo{year}{2020}{\natexlab{a}}) \bibinfo{pages}{234--247}.
  \DOIprefix\doi{10.1016/j.neunet.2020.05.013}.
\bibitem[{Hart et~al.(2020{\natexlab{b}})Hart, Hook, and Dawes}]{hart2020echo}
\bibinfo{author}{A.~G. Hart}, \bibinfo{author}{J.~L. Hook},
  \bibinfo{author}{J.~H. Dawes},
\newblock \bibinfo{title}{Echo state networks trained by tikhonov least squares
  are l2($\mu$) approximators of ergodic dynamical systems},
\newblock \bibinfo{journal}{arXiv preprint arXiv:2005.06967}
  (\bibinfo{year}{2020}{\natexlab{b}}).
\bibitem[{Gonon et~al.(2020)Gonon, Grigoryeva, and Ortega}]{gonon2020memory}
\bibinfo{author}{L.~Gonon}, \bibinfo{author}{L.~Grigoryeva},
  \bibinfo{author}{J.-P. Ortega},
\newblock \bibinfo{title}{Memory and forecasting capacities of nonlinear
  recurrent networks},
\newblock \bibinfo{journal}{Physica D: Nonlinear Phenomena}
  \bibinfo{volume}{414} (\bibinfo{year}{2020}) \bibinfo{pages}{132721}.
  \DOIprefix\doi{10.1016/j.physd.2020.132721}.
\bibitem[{Massar and Massar(2013)}]{massar2013mean}
\bibinfo{author}{M.~Massar}, \bibinfo{author}{S.~Massar},
\newblock \bibinfo{title}{Mean-field theory of echo state networks},
\newblock \bibinfo{journal}{Physical Review E} \bibinfo{volume}{87}
  (\bibinfo{year}{2013}) \bibinfo{pages}{042809}.
\bibitem[{Mastrogiuseppe and Ostojic(2019)}]{mastrogiuseppe2019geometrical}
\bibinfo{author}{F.~Mastrogiuseppe}, \bibinfo{author}{S.~Ostojic},
\newblock \bibinfo{title}{A geometrical analysis of global stability in trained
  feedback networks},
\newblock \bibinfo{journal}{Neural Computation} \bibinfo{volume}{31}
  (\bibinfo{year}{2019}) \bibinfo{pages}{1139--1182}.
\bibitem[{Rivkind and Barak(2017)}]{rivkind2017local}
\bibinfo{author}{A.~Rivkind}, \bibinfo{author}{O.~Barak},
\newblock \bibinfo{title}{Local dynamics in trained recurrent neural networks},
\newblock \bibinfo{journal}{Physical Review Letters} \bibinfo{volume}{118}
  (\bibinfo{year}{2017}) \bibinfo{pages}{258101}.
\bibitem[{Verstraeten et~al.(2010)Verstraeten, Dambre, Dutoit, and
  Schrauwen}]{verstraeten2010memory}
\bibinfo{author}{D.~Verstraeten}, \bibinfo{author}{J.~Dambre},
  \bibinfo{author}{X.~Dutoit}, \bibinfo{author}{B.~Schrauwen},
\newblock \bibinfo{title}{Memory versus non-linearity in reservoirs},
\newblock in: \bibinfo{booktitle}{The 2010 international joint conference on
  neural networks (IJCNN)}, \bibinfo{organization}{IEEE}, \bibinfo{year}{2010},
  pp. \bibinfo{pages}{1--8}.
\bibitem[{Goudarzi et~al.(2016)Goudarzi, Marzen, Banda, Feldman, Teuscher, and
  Stefanovic}]{goudarzi2016memory}
\bibinfo{author}{A.~Goudarzi}, \bibinfo{author}{S.~Marzen},
  \bibinfo{author}{P.~Banda}, \bibinfo{author}{G.~Feldman},
  \bibinfo{author}{C.~Teuscher}, \bibinfo{author}{D.~Stefanovic},
\newblock \bibinfo{title}{Memory and information processing in recurrent neural
  networks},
\newblock \bibinfo{journal}{arXiv preprint arXiv:1604.06929}
  (\bibinfo{year}{2016}).
\bibitem[{Marzen(2017)}]{marzen2017difference}
\bibinfo{author}{S.~Marzen},
\newblock \bibinfo{title}{Difference between memory and prediction in linear
  recurrent networks},
\newblock \bibinfo{journal}{Physical Review E} \bibinfo{volume}{96}
  (\bibinfo{year}{2017}) \bibinfo{pages}{032308}.
  \DOIprefix\doi{10.1103/PhysRevE.96.032308}.
\bibitem[{Ti{\v{n}}o(2020)}]{tino2020dynamical}
\bibinfo{author}{P.~Ti{\v{n}}o},
\newblock \bibinfo{title}{Dynamical systems as temporal feature spaces.},
\newblock \bibinfo{journal}{Journal of Machine Learning Research}
  \bibinfo{volume}{21} (\bibinfo{year}{2020}) \bibinfo{pages}{1--42}.
\bibitem[{Verzelli et~al.(2020)Verzelli, Alippi, Livi, and
  Tino}]{verzelli2020input}
\bibinfo{author}{P.~Verzelli}, \bibinfo{author}{C.~Alippi},
  \bibinfo{author}{L.~Livi}, \bibinfo{author}{P.~Tino},
\newblock \bibinfo{title}{Input representation in recurrent neural networks
  dynamics},
\newblock \bibinfo{journal}{arXiv preprint arXiv:2003.10585}
  (\bibinfo{year}{2020}).
\bibitem[{Ganguli et~al.(2008)Ganguli, Huh, and
  Sompolinsky}]{ganguli2008memory}
\bibinfo{author}{S.~Ganguli}, \bibinfo{author}{D.~Huh},
  \bibinfo{author}{H.~Sompolinsky},
\newblock \bibinfo{title}{Memory traces in dynamical systems},
\newblock \bibinfo{journal}{Proceedings of the National Academy of Sciences}
  \bibinfo{volume}{105} (\bibinfo{year}{2008}) \bibinfo{pages}{18970--18975}.
  \DOIprefix\doi{10.1073/pnas.0804451105}.
\bibitem[{Tanaka et~al.(2019)Tanaka, Yamane, Héroux, Nakane, Kanazawa, Takeda,
  Numata, Nakano, and Hirose}]{TANAKA2019100}
\bibinfo{author}{G.~Tanaka}, \bibinfo{author}{T.~Yamane},
  \bibinfo{author}{J.~B. Héroux}, \bibinfo{author}{R.~Nakane},
  \bibinfo{author}{N.~Kanazawa}, \bibinfo{author}{S.~Takeda},
  \bibinfo{author}{H.~Numata}, \bibinfo{author}{D.~Nakano},
  \bibinfo{author}{A.~Hirose},
\newblock \bibinfo{title}{Recent advances in physical reservoir computing: {A}
  review},
\newblock \bibinfo{journal}{Neural Networks} \bibinfo{volume}{115}
  (\bibinfo{year}{2019}) \bibinfo{pages}{100 -- 123}.
  \DOIprefix\doi{10.1016/j.neunet.2019.03.005}.
\bibitem[{Yildiz et~al.(2012)Yildiz, Jaeger, and Kiebel}]{yildiz2012re}
\bibinfo{author}{I.~B. Yildiz}, \bibinfo{author}{H.~Jaeger},
  \bibinfo{author}{S.~J. Kiebel},
\newblock \bibinfo{title}{Re-visiting the echo state property},
\newblock \bibinfo{journal}{Neural Networks} \bibinfo{volume}{35}
  (\bibinfo{year}{2012}) \bibinfo{pages}{1--9}.
\bibitem[{Zhang et~al.(2011)Zhang, Miller, and Wang}]{zhang2011nonlinear}
\bibinfo{author}{B.~Zhang}, \bibinfo{author}{D.~J. Miller},
  \bibinfo{author}{Y.~Wang},
\newblock \bibinfo{title}{Nonlinear system modeling with random matrices: echo
  state networks revisited},
\newblock \bibinfo{journal}{IEEE Transactions on Neural Networks and Learning
  Systems} \bibinfo{volume}{23} (\bibinfo{year}{2011})
  \bibinfo{pages}{175--182}.
\bibitem[{Basterrech(2017)}]{basterrech2017empirical}
\bibinfo{author}{S.~Basterrech},
\newblock \bibinfo{title}{Empirical analysis of the necessary and sufficient
  conditions of the echo state property},
\newblock in: \bibinfo{booktitle}{2017 International Joint Conference on Neural
  Networks (IJCNN)}, \bibinfo{organization}{IEEE}, \bibinfo{year}{2017}, pp.
  \bibinfo{pages}{888--896}.
\bibitem[{Manjunath and Jaeger(2013)}]{manjunath2013echo}
\bibinfo{author}{G.~Manjunath}, \bibinfo{author}{H.~Jaeger},
\newblock \bibinfo{title}{Echo state property linked to an input: Exploring a
  fundamental characteristic of recurrent neural networks},
\newblock \bibinfo{journal}{Neural Computation} \bibinfo{volume}{25}
  (\bibinfo{year}{2013}) \bibinfo{pages}{671--696}.
\bibitem[{Caluwaerts et~al.(2013)Caluwaerts, Wyffels, Dieleman, and
  Schrauwen}]{caluwaerts2013spectral}
\bibinfo{author}{K.~Caluwaerts}, \bibinfo{author}{F.~Wyffels},
  \bibinfo{author}{S.~Dieleman}, \bibinfo{author}{B.~Schrauwen},
\newblock \bibinfo{title}{The spectral radius remains a valid indicator of the
  echo state property for large reservoirs},
\newblock in: \bibinfo{booktitle}{The 2013 International Joint Conference on
  Neural Networks (IJCNN)}, \bibinfo{organization}{IEEE}, \bibinfo{year}{2013},
  pp. \bibinfo{pages}{1--6}.
\bibitem[{Ceni et~al.(2020)Ceni, Ashwin, Livi, and
  Postlethwaite}]{ceni2020nESP}
\bibinfo{author}{A.~Ceni}, \bibinfo{author}{P.~Ashwin},
  \bibinfo{author}{L.~Livi}, \bibinfo{author}{C.~Postlethwaite},
\newblock \bibinfo{title}{The echo index and multistability in input-driven
  recurrent neural networks},
\newblock \bibinfo{journal}{Physica D} \bibinfo{volume}{412}
  (\bibinfo{year}{2020}). \DOIprefix\doi{10.1016/j.physd.2020.132609}.
\bibitem[{Lu and Bassett(2020)}]{lu2020invertible}
\bibinfo{author}{Z.~Lu}, \bibinfo{author}{D.~S. Bassett},
\newblock \bibinfo{title}{Invertible generalized synchronization: A putative
  mechanism for implicit learning in neural systems},
\newblock \bibinfo{journal}{Chaos: An Interdisciplinary Journal of Nonlinear
  Science} \bibinfo{volume}{30} (\bibinfo{year}{2020}) \bibinfo{pages}{063133}.
\bibitem[{Weng et~al.(2019)Weng, Yang, Gu, Zhang, and
  Small}]{weng2019synchronization}
\bibinfo{author}{T.~Weng}, \bibinfo{author}{H.~Yang}, \bibinfo{author}{C.~Gu},
  \bibinfo{author}{J.~Zhang}, \bibinfo{author}{M.~Small},
\newblock \bibinfo{title}{Synchronization of chaotic systems and their
  machine-learning models},
\newblock \bibinfo{journal}{Physical Review E} \bibinfo{volume}{99}
  (\bibinfo{year}{2019}) \bibinfo{pages}{042203}.
\bibitem[{Lymburn et~al.(2019)Lymburn, Walker, Small, and
  J{\"u}ngling}]{lymburn2019reservoir}
\bibinfo{author}{T.~Lymburn}, \bibinfo{author}{D.~M. Walker},
  \bibinfo{author}{M.~Small}, \bibinfo{author}{T.~J{\"u}ngling},
\newblock \bibinfo{title}{The reservoir’s perspective on generalized
  synchronization},
\newblock \bibinfo{journal}{Chaos: An Interdisciplinary Journal of Nonlinear
  Science} \bibinfo{volume}{29} (\bibinfo{year}{2019}) \bibinfo{pages}{093133}.
\bibitem[{Grigoryeva et~al.(2020)Grigoryeva, Hart, and
  Ortega}]{grigoryeva2020chaos}
\bibinfo{author}{L.~Grigoryeva}, \bibinfo{author}{A.~Hart},
  \bibinfo{author}{J.-P. Ortega},
\newblock \bibinfo{title}{Chaos on compact manifolds: Differentiable
  synchronizations beyond takens},
\newblock \bibinfo{journal}{arXiv preprint arXiv:2010.03218}
  (\bibinfo{year}{2020}).
\bibitem[{Afraimovich et~al.(1986)Afraimovich, Verichev, and
  Rabinovich}]{afraimovich1986stochastic}
\bibinfo{author}{V.~Afraimovich}, \bibinfo{author}{N.~Verichev},
  \bibinfo{author}{M.~I. Rabinovich},
\newblock \bibinfo{title}{Stochastic synchronization of oscillation in
  dissipative systems},
\newblock \bibinfo{journal}{Radiophysics and Quantum Electronics}
  \bibinfo{volume}{29} (\bibinfo{year}{1986}) \bibinfo{pages}{795--803}.
\bibitem[{Rulkov et~al.(1995)Rulkov, Sushchik, Tsimring, and
  Abarbanel}]{rulkov1995generalized}
\bibinfo{author}{N.~F. Rulkov}, \bibinfo{author}{M.~M. Sushchik},
  \bibinfo{author}{L.~S. Tsimring}, \bibinfo{author}{H.~D. Abarbanel},
\newblock \bibinfo{title}{Generalized synchronization of chaos in directionally
  coupled chaotic systems},
\newblock \bibinfo{journal}{Physical Review E} \bibinfo{volume}{51}
  (\bibinfo{year}{1995}) \bibinfo{pages}{980}.
\bibitem[{Pecora et~al.(1997)Pecora, Carroll, Johnson, Mar, and
  Heagy}]{pecora1997fundamentals}
\bibinfo{author}{L.~M. Pecora}, \bibinfo{author}{T.~L. Carroll},
  \bibinfo{author}{G.~A. Johnson}, \bibinfo{author}{D.~J. Mar},
  \bibinfo{author}{J.~F. Heagy},
\newblock \bibinfo{title}{Fundamentals of synchronization in chaotic systems,
  concepts, and applications},
\newblock \bibinfo{journal}{Chaos: An Interdisciplinary Journal of Nonlinear
  Science} \bibinfo{volume}{7} (\bibinfo{year}{1997})
  \bibinfo{pages}{520--543}.
\bibitem[{Parlitz(2012)}]{parlitz2012detecting}
\bibinfo{author}{U.~Parlitz},
\newblock \bibinfo{title}{Detecting generalized synchronization},
\newblock \bibinfo{journal}{Nonlinear Theory and Its Applications, IEICE}
  \bibinfo{volume}{3} (\bibinfo{year}{2012}) \bibinfo{pages}{113--127}.
\bibitem[{Boccaletti et~al.(2002)Boccaletti, Kurths, Osipov, Valladares, and
  Zhou}]{boccaletti2002synchronization}
\bibinfo{author}{S.~Boccaletti}, \bibinfo{author}{J.~Kurths},
  \bibinfo{author}{G.~Osipov}, \bibinfo{author}{D.~Valladares},
  \bibinfo{author}{C.~Zhou},
\newblock \bibinfo{title}{The synchronization of chaotic systems},
\newblock \bibinfo{journal}{Physics Reports} \bibinfo{volume}{366}
  (\bibinfo{year}{2002}) \bibinfo{pages}{1--101}.
\bibitem[{Manjunath et~al.(2012)Manjunath, Tino, and
  Jaeger}]{manjunath2012theory}
\bibinfo{author}{G.~Manjunath}, \bibinfo{author}{P.~Tino},
  \bibinfo{author}{H.~Jaeger},
\newblock \bibinfo{title}{Theory of input driven dynamical systems},
\newblock \bibinfo{journal}{dice. ucl. ac. be, number April}
  (\bibinfo{year}{2012}) \bibinfo{pages}{25--27}.
\bibitem[{Pyragas(1996)}]{pyragas1996weak}
\bibinfo{author}{K.~Pyragas},
\newblock \bibinfo{title}{Weak and strong synchronization of chaos},
\newblock \bibinfo{journal}{Physical Review E} \bibinfo{volume}{54}
  (\bibinfo{year}{1996}) \bibinfo{pages}{R4508}.
\bibitem[{Takens(1981)}]{takens1981detecting}
\bibinfo{author}{F.~Takens},
\newblock \bibinfo{title}{Detecting strange attractors in turbulence},
\newblock in: \bibinfo{booktitle}{Dynamical Systems and Turbulence},
  \bibinfo{publisher}{Springer}, \bibinfo{year}{1981}, pp.
  \bibinfo{pages}{366--381}.
\bibitem[{Shawe-Taylor and Cristianini(2004)}]{shawetaylor+cristianini2004}
\bibinfo{author}{J.~Shawe-Taylor}, \bibinfo{author}{N.~Cristianini},
  \bibinfo{title}{{Kernel Methods for Pattern Analysis}},
  \bibinfo{publisher}{Cambridge University Press}, \bibinfo{address}{Cambridge,
  UK}, \bibinfo{year}{2004}.
\bibitem[{Shi and Han(2007)}]{shi2007support}
\bibinfo{author}{Z.~Shi}, \bibinfo{author}{M.~Han},
\newblock \bibinfo{title}{Support vector echo-state machine for chaotic
  time-series prediction},
\newblock \bibinfo{journal}{IEEE Transactions on Neural Networks}
  \bibinfo{volume}{18} (\bibinfo{year}{2007}) \bibinfo{pages}{359--372}.
\bibitem[{Shalev-Shwartz and Ben-David(2014)}]{shalev2014understanding}
\bibinfo{author}{S.~Shalev-Shwartz}, \bibinfo{author}{S.~Ben-David},
  \bibinfo{title}{Understanding machine learning: From theory to algorithms},
  \bibinfo{publisher}{Cambridge university press}, \bibinfo{year}{2014}.
\bibitem[{Birkhoff(1931)}]{birkhoff1931proof}
\bibinfo{author}{G.~D. Birkhoff},
\newblock \bibinfo{title}{Proof of the ergodic theorem},
\newblock \bibinfo{journal}{Proceedings of the National Academy of Sciences}
  \bibinfo{volume}{17} (\bibinfo{year}{1931}) \bibinfo{pages}{656--660}.
\bibitem[{Platt et~al.(2021)Platt, Wong, Clark, Penny, and
  Abarbanel}]{platt2021forecasting}
\bibinfo{author}{J.~A. Platt}, \bibinfo{author}{A.~S. Wong},
  \bibinfo{author}{R.~Clark}, \bibinfo{author}{S.~G. Penny},
  \bibinfo{author}{H.~D. Abarbanel},
\newblock \bibinfo{title}{Forecasting using reservoir computing: The role of
  generalized synchronization},
\newblock \bibinfo{journal}{arXiv preprint arXiv:2103.00362}
  (\bibinfo{year}{2021}).
\bibitem[{Gallicchio et~al.(2017)Gallicchio, Micheli, and
  Pedrelli}]{gallicchio2017deep}
\bibinfo{author}{C.~Gallicchio}, \bibinfo{author}{A.~Micheli},
  \bibinfo{author}{L.~Pedrelli},
\newblock \bibinfo{title}{Deep reservoir computing: A critical experimental
  analysis},
\newblock \bibinfo{journal}{Neurocomputing} \bibinfo{volume}{268}
  (\bibinfo{year}{2017}) \bibinfo{pages}{87--99}.
\bibitem[{L{\o}kse et~al.(2017)L{\o}kse, Bianchi, and
  Jenssen}]{lokse2017training}
\bibinfo{author}{S.~L{\o}kse}, \bibinfo{author}{F.~M. Bianchi},
  \bibinfo{author}{R.~Jenssen},
\newblock \bibinfo{title}{Training echo state networks with regularization
  through dimensionality reduction},
\newblock \bibinfo{journal}{Cognitive Computation} \bibinfo{volume}{9}
  (\bibinfo{year}{2017}) \bibinfo{pages}{364--378}.
\bibitem[{Ott(2002)}]{ott2002chaos}
\bibinfo{author}{E.~Ott}, \bibinfo{title}{Chaos in dynamical systems},
  \bibinfo{publisher}{Cambridge university press}, \bibinfo{year}{2002}.
\bibitem[{Pecora and Carroll(1990)}]{pecora1990synchronization}
\bibinfo{author}{L.~M. Pecora}, \bibinfo{author}{T.~L. Carroll},
\newblock \bibinfo{title}{Synchronization in chaotic systems},
\newblock \bibinfo{journal}{Physical Review Letters} \bibinfo{volume}{64}
  (\bibinfo{year}{1990}) \bibinfo{pages}{821}.
\bibitem[{Kocarev and Parlitz(1996)}]{kocarev1996generalized}
\bibinfo{author}{L.~Kocarev}, \bibinfo{author}{U.~Parlitz},
\newblock \bibinfo{title}{Generalized synchronization, predictability, and
  equivalence of unidirectionally coupled dynamical systems},
\newblock \bibinfo{journal}{Physical Review Letters} \bibinfo{volume}{76}
  (\bibinfo{year}{1996}) \bibinfo{pages}{1816}.
\bibitem[{Lorenz(1963)}]{lorenz1963deterministic}
\bibinfo{author}{E.~N. Lorenz},
\newblock \bibinfo{title}{Deterministic nonperiodic flow},
\newblock \bibinfo{journal}{Journal of the Atmospheric Sciences}
  \bibinfo{volume}{20} (\bibinfo{year}{1963}) \bibinfo{pages}{130--141}.
\bibitem[{R{\"o}ssler(1976)}]{rossler1976equation}
\bibinfo{author}{O.~E. R{\"o}ssler},
\newblock \bibinfo{title}{An equation for continuous chaos},
\newblock \bibinfo{journal}{Physics Letters A} \bibinfo{volume}{57}
  (\bibinfo{year}{1976}) \bibinfo{pages}{397--398}.

\end{thebibliography}

\appendix

\section{Synchronization of identical systems} \label{sec:identical_sync}

Following \cite{ott2002chaos}, we start by recalling the concept of \emph{sensitive dependence on initial conditions}. 
Consider two identical $d$-dimensional chaotic systems, say $a$ and $b$, described by:
\begin{subequations}\label{eqn:system}
\begin{align} 
    \vec{x}_a(t+\tau) &= \vec{F}(\vec{x}_a(t)) \\
    \vec{x}_b(t+\tau) &= \vec{F}(\vec{x}_b(t))
\end{align}{}
\end{subequations}
where the function $\vec{F}$ is the same for both systems.
If the initial conditions differ even slightly, then the chaotic nature of the system will lead to exponential divergence: the two systems posses the same attractor but their motion will be uncorrelated over time.

In this context, an instance of chaos synchronization consists of designing a coupling between the two systems such that the two trajectories, $\vec{x}_a(t)$ and $\vec{x}_b(t)$, become identical asymptotically with time. 
That is, if $\vec{x}_a(t) \approx \vec{x}_b(t)$ then $\lVert\vec{x}_a(t) - \vec{x}_b(t)\rVert \to 0$ as $t \to \infty$.
A possible coupling for \eqref{eqn:system} might be:
\begin{subequations}\label{eqn:coupling}
\begin{align} 
    \vec{x}_a(t+\tau) &= \vec{F}(\vec{x}_a(t)) + \vec{c}_a \left(\vec{x}_a(t) -\vec{x}_b(t) \right)\\
    \vec{x}_b(t+\tau) &= \vec{F}(\vec{x}_b(t)) + \vec{c}_b \left(\vec{x}_b(t)  - \vec{x}_a(t)\right)
\end{align}
\end{subequations}

The $\vec{c}_a= [c_{a,1}, c_{a,2}, \dots c_{a,d}]$ and $\vec{c}_b= [c_{b,1}, c_{b,2}, \dots c_{b,d}]$ are the \emph{coupling constants}. If all the $c_a$'s are null, we say that there is one-way coupling from $a$ to $b$, since the state of $a$ influences $b$ but $b$ does no influence $a$. If $c_{a,i} \ne 0 $ and $c_{b,i} \ne 0 $ for at least one $i$, we say that there is a two-way coupling. 

System in \eqref{eqn:coupling} is, as a whole, a $2d$-dimensional dynamical system resulting from the coupling of the two original systems.
Note that if synchronization is achieved, $\vec{x}_a(t) = \vec{x}_b(t)$: this means that the coupling terms are null.

In the $2d$-dimensional state-space of system \eqref{eqn:coupling}, the synchronized state $\vec{x}_a = \vec{x}_b$ represents an $d$-dimensional invariant manifold.
On this manifold, \eqref{eqn:coupling} reduces to \eqref{eqn:system}.

\section{Complete synchronization and asymptotic stability} \label{sec:complet_synch}

In the framework introduced for system \ref{eqn:drive_response}, we now introduce a \emph{driven replica subsystem}:
\begin{equation}\label{eqn:driven_replica}
    \tilde{\vec{r}}(t+\tau) = \vec{f}(\tilde{\vec{r}}(t), \vec{h}(\vec{s}(t+\tau)))
\end{equation}
Note that $\vec{f}$ is the same as in \eqref{eqn:response}. We then take the sequence of states $\vec{s}(t)$ from \eqref{eqn:drive} and use $\vec{h}(\vec{s})$ to feed the replica subsystem \eqref{eqn:driven_replica}.
The complete synchronization \cite{pecora1990synchronization} between the response \eqref{eqn:response} and its replica \eqref{eqn:driven_replica} is defined as the identity of the trajectories of $\vec{r}$ and $\tilde{\vec{r}}$.
In more formal terms, we are requiring the \emph{asymptotic stability} of the response with respect to the replica subsystem \cite[Sec.~3.6]{boccaletti2002synchronization}.
\begin{defn}[Asymptotic stability]\label{def:asymptotic_stability}
A dynamical system is said to be \emph{asymptotically stable} if, for any two copies $\vec{r}$ and $\tilde{\vec{r}}$ of the system driven by the same input $\vec{u}(t)$ and starting from different initial conditions in $B_{r}$, it holds that
\begin{equation}
\label{eqn:asymp_stability_synch}
\lim_{t \to \infty } \lVert \vec{r}(t, \vec{u}(t+\tau))  - \tilde{\vec{r}}(t, \vec{u}(t+\tau))  \rVert = 0
\end{equation}
\end{defn}{}

In our case, $\vec{u}(t) = \vec{h}(\vec{s(t)})$.
The state of the full dynamical system is now constituted by \eqref{eqn:drive_response} and \eqref{eqn:driven_replica}, and thus it is $d_s + 2d_r$ dimensional. 
The synchronized state $\hat{\vec{r}} = \vec{r}$ represents an ($d_s + d_r$)-dimensional manifold embedded in the state-space of the full system.

In \cite{kocarev1996generalized} the authors proved a necessary and sufficient condition for the \ac{GS} between the driver $\vec{s}$ (through $\vec{u} = \vec{h}(\vec{s})$) and the response $\vec{r}$ to hold: \ac{GS} occurs if and only if, for all initial conditions in $\mathcal{B}$, the response system is asymptotically stable.

\section{Lorenz System}\label{sec:lorenz}

The Lorenz system \cite{lorenz1963deterministic} is a $3$-dimensional dynamical system characterizing a simple model for atmospheric convection.
Its equations read:
\begin{align}
\begin{split}
  \dot{x} &= \sigma(y - x)\\
  \dot{y} &= (\rho-z) x - y\\
  \dot{z} &= x y - \beta z
\end{split}
\end{align}

where $x = x(t)$, $y = y(t)$, $z = z(t)$ are the variables, $\sigma$, $\rho$ and $\beta$ are the  model parameters and the dot denotes the first-order derivative with respect to time $t$.
In this work, we choose the commonly used values $\sigma = 10$, $\rho=28$ and $\beta=8/3$, for which the system is known to be a chaotic one and to have a strange attractor.

\section{R{\"o}ssler System}\label{sec:roessler}

The R{\"o}ssler system \cite{rossler1976equation} is a $3$-dimensional chaotic dynamical system defined as follows:
\begin{align}
\begin{split}
  \dot{x} &= -y -z \\
  \dot{y} &= x + a y\\
  \dot{z} &= b + z(x-c)
\end{split}
\end{align}
where $x = x(t)$, $y = y(t)$, $z = z(t)$ are the variables and $a$, $b$, and $c$ are the model parameters, which in our paper are set to $a= 0.1$, $b=0.1$, and $c=14$. The dot denotes the first-order derivative with respect to time $t$.

\section{Measurement noise} \label{sec:noise}

In many real situations the input is corrupted by some noise, so that instead of reading just $\vec{u}(t)$ one actually reads $\vec{u}(t) + \vec{\epsilon}(t)$, $\vec{\epsilon}(t)$ being i.i.d. noise.
This lead to the following state-update for the reservoir:
\begin{align}
    \vec{r}(t+1) &= \vec{f}(\vec{r}(t), \vec{u}(t) + \vec{\epsilon}(t)) \\
    \nonumber&\approx \vec{f}(\vec{r}(t), \vec{u}(t)) + \vec{f}'(\vec{r}(t), \vec{u}(t)) \vec{\epsilon}(t)
\end{align}

This will affect the reservoir dynamics in general, but when the noise is small we can still hope that the trajectory will not be too far from the one generated without noise. That is, we assume it is possible to write each point as $\vec{r}(t) + \vec{\eta}(t)$.
This will be in fact guaranteed by the \ac{GS}, which requires the synchronization manifold not only to exist, but also to be attractive \cite{kocarev1996generalized}.
Note that $\vec{\eta}(t)$ is not i.i.d. anymore.

The synchronization problem (w.r.t. to the \emph{true} system state) becomes:
\begin{equation}
    \vec{s} = \vec{\phi} ( \vec{r} + \vec{\eta}  )
\end{equation}

We can make use of the smoothness of $\vec{\phi}$ to write a first-order approximation of the source system state as follows:
\begin{equation}
    \vec{s} \approx \vec{\phi} ( \vec{r} ) + \vec{\phi}'( \vec{r} ) \vec{\eta}  
\end{equation}

Such an approximation allows us to introduce a measure of \emph{synchronization error} due to noise, which reads:
\begin{equation}
    E_n := \lVert \vec{s} - \vec{\phi} ( \vec{r} ) \rVert  \approx 
    \lVert\vec{\phi}' \vec{\eta}  \rVert \le
    \lVert\vec{\phi}'\rVert   \lVert\vec{\eta}  \rVert
\end{equation}

For the observer task (see \ref{subsec:Reservoir_Observer}), in the common case of a linear readout, $\vec{\phi}'$ is simply the pseudo-inverse of the readout matrix $\matr{W}^{*}_\text{out}$, whose singular values are the reciprocal of the singular values of $\matr{W}_\text{out}$. This implies the following bound on the synchronization error due to noise,
\begin{equation}
   E_n \le \lVert \matr{W}^{*}_\text{out} \rVert  \lVert\vec{\eta}  \rVert
   = \frac{\lVert\vec{\eta} \rVert} { \min_i \sigma_i(\matr{W}_\text{out})}
\end{equation}
where $\sigma_i(\matr{W}_\text{out})$ denote the non-null singular values of $\matr{W}_\text{out}$.

\begin{acronym}
    \acro{bESN}{binary ESN}
    \acro{CH}{Cayley-Hamilton}
    \acro{CS}{Complete Synchronization}
	\acro{EoC}{Edge of Criticality}
	\acro{ESN}{Echo State Network}
	\acro{ESP}{Echo State Property}
	\acro{FIM}{Fisher Information Matrix}
	\acro{FPM}{Fractal Predicting Machine}
	\acro{GS}{Generalized Synchronization}
	\acro{LLE}{Local Lyapunov Exponent}
	\acro{LSM}{Liquid State Machine}
	\acro{MFNN}{Mutual False Nearest Neighbors}
	\acro{MFT}{Mean Field Theory}
	\acro{MSV}{Maximum Singular Value}
	\acro{ML}{Machine Learning}
	\acro{MSE}{Mean Squared Error}
	\acro{MSO}{Multiple Superimposed Oscillator}
	\acro{NRMSE}{Normalized Root Mean Squared Error}
	\acro{PDF}{Probability Density Function}
	\acro{RC}{Reservoir Computing}
	\acro{RCN}{Reservoir Computing Network}
	\acro{RBN}{Random Boolean Network}
	\acro{RMSE}{Root Mean Squared Error}
	\acro{RNN}{Recurrent Neural Network}
	\acro{RP}{Recurrency Plot}
	\acro{SI}{Supporting Information}
	\acro{SR}{Spectral Radius}
\end{acronym}

\end{document}